\documentclass{article}

\usepackage[final]{neurips}

\usepackage{microtype}
\usepackage{graphicx}
\usepackage{subfigure}
\usepackage{booktabs} 
\usepackage{hyperref}
\usepackage{placeins}
\usepackage{algorithm}
\usepackage{algpseudocode}
\usepackage[utf8]{inputenc} 
\usepackage[T1]{fontenc}    
\usepackage{microtype}      

\usepackage{amsfonts}       
\usepackage{mathtools}      
\usepackage{nicefrac}       

\usepackage[table]{xcolor}   

\usepackage{tikz}           
\usetikzlibrary{positioning}
\usepackage{pgfplots}       
\pgfplotsset{compat=1.17}    
\usepgfplotslibrary{groupplots}
\usepackage{booktabs}       

\usepackage{enumitem}       
\usepackage{wrapfig}        

\usepackage{hyperref}       

\usepackage{amsmath}
\usepackage{amssymb}
\usepackage{mathtools}
\usepackage{amsthm}
\usepackage{enumitem}
\usepackage[framemethod=tikz]{mdframed}
\usepackage{physics}

\usepackage[capitalize,noabbrev]{cleveref}
\newcolumntype{C}{>{\centering\arraybackslash}p{0.62cm}}

\definecolor{lblue}{HTML}{45b6fe}
\definecolor{lred}{HTML}{FF474C}

\theoremstyle{plain}
\newtheorem{theorem}{Theorem}[section]

\newtheorem{remark}[theorem]{Remark}

\title{Energy Matching: Unifying Flow Matching and Energy-Based Models for Generative Modeling}

\author{
  \textbf{Michal Balcerak}\\
  University of Zurich\\
  \texttt{michal.balcerak@uzh.ch}
  \And
  \textbf{Tamaz Amiranashvili}\\
  University of Zurich\\
  Technical University of Munich \\
  \And
  \textbf{Antonio Terpin}\\
  ETH Zurich\\
  \And
  \textbf{Suprosanna Shit}\\
  University of Zurich\\
  \And
  \textbf{Lea Bogensperger}\\
  University of Zurich\\
  \And
  \textbf{Sebastian Kaltenbach}\\
  Harvard University\\
  \And
  \textbf{Petros Koumoutsakos}\\ 
  Harvard University\\
  \And
  \textbf{Bjoern Menze}\\
  University of Zurich\\
}

\usepackage[acronym]{glossaries}
\newacronym[plural=EBMs, firstplural=energy-based models (EBMs)]{ebm}{EBM}{energy-based model}
\newacronym{ot}{OT}{optimal transport}
\newacronym{lid}{LID}{local intrinsic dimension}
\newacronym{mcmc}{MCMC}{Markov chain Monte Carlo}
\newacronym{jko}{JKO}{Jordan–Kinderlehrer–Otto}
\newacronym{fid}{FID}{Fr\'echet inception distance}
\DeclareMathOperator{\Lot}{\mathcal{L}_{\mathrm{OT}}}
\DeclareMathOperator{\Lcd}{\mathcal{L}_{\mathrm{CD}}}
\DeclareMathOperator{\Tot}{\tau^\ast}
\DeclareMathOperator{\samplingTime}{\tau_s}
\begin{document}

\maketitle

\begin{abstract}
Current state-of-the-art generative models map noise to data distributions by matching flows or scores. A key limitation of these models is their inability to readily integrate available partial observations and additional priors. 
In contrast, \glspl*{ebm} address this by incorporating corresponding scalar energy terms. Here, we propose \textit{Energy Matching}, a framework that endows flow-based approaches with the flexibility of \glspl*{ebm}. Far from the data manifold, samples move from noise to data along irrotational, optimal transport paths. As they approach the data manifold, an entropic energy term guides the system into a Boltzmann equilibrium distribution, explicitly capturing the underlying likelihood structure of the data. We parameterize these dynamics with a single time-independent scalar field, which serves as both a powerful generator and a flexible prior for effective regularization of inverse problems. The present  method substantially outperforms existing \glspl*{ebm} on CIFAR-10 and ImageNet generation in terms of fidelity, while retaining simulation-free training of transport-based approaches away from the data manifold. Furthermore, we leverage the flexibility of the method to introduce an interaction energy that supports the exploration of diverse modes, which we demonstrate in a controlled protein generation setting. This approach learns a scalar potential energy, without time conditioning, auxiliary generators, or additional networks, marking  a significant departure from recent \gls *{ebm} methods. We believe this simplified yet rigorous formulation significantly advances \glspl*{ebm} capabilities and paves the way for their wider adoption in generative modeling in diverse domains.
\end{abstract}


\section{Introduction}
\label{sec:intro}
\begin{figure}
    \centering
\begin{tikzpicture}[>=stealth, node distance=0cm]
    \def\skiptitle{1.9}   
    \def\skipdesc{2.0}    
    \def\skipimage{3.5}
    \def\imagewidth{3.3}

    \node (imgA)
    {\includegraphics[width=\imagewidth cm]{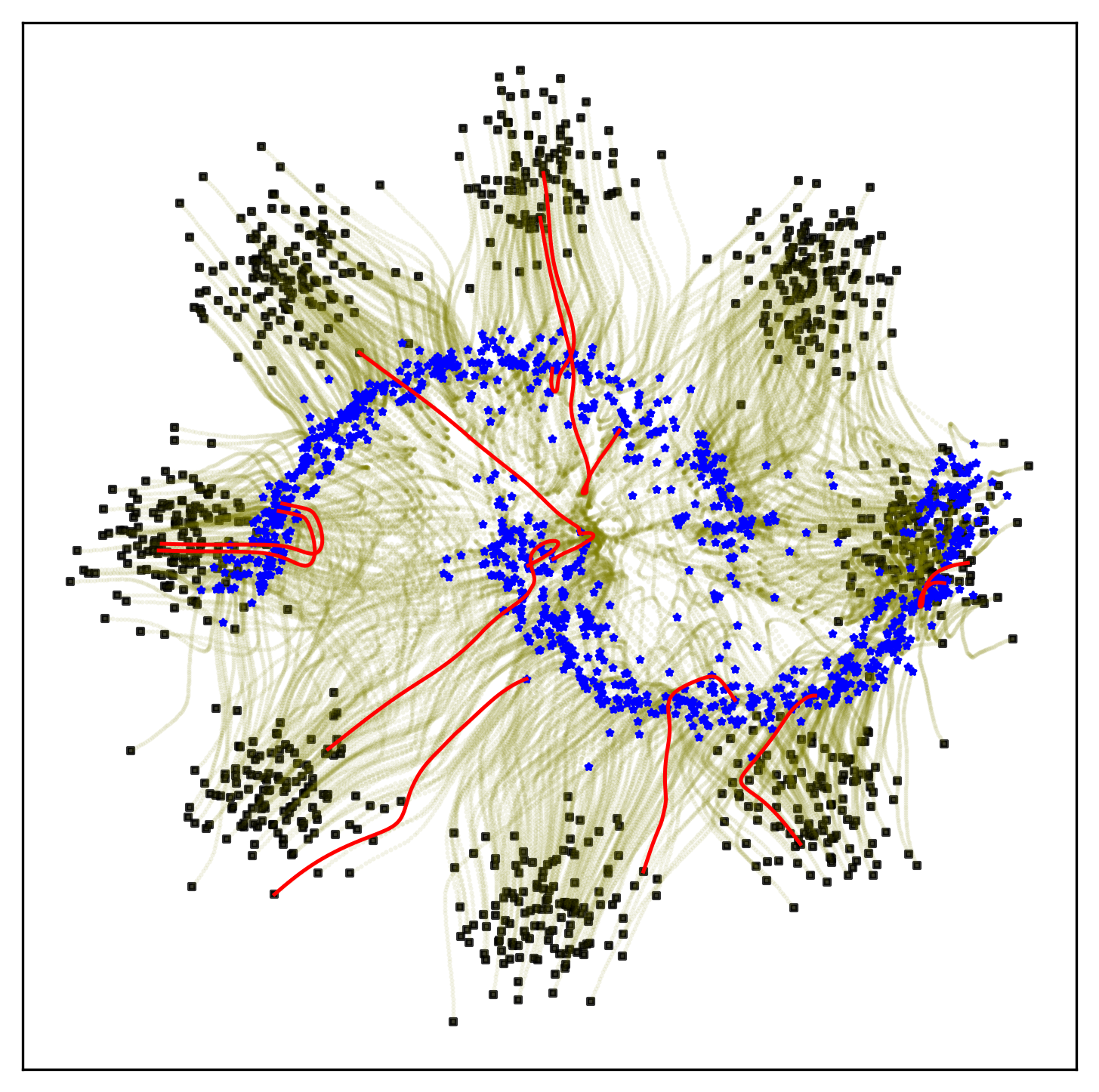}};
    \node[anchor=south, above of=imgA, node distance=\skiptitle cm]
         {Action Matching};
    \node[anchor=north, below of=imgA, node distance=\skipdesc cm]
         {$s_{\theta}(x, t) \in \mathbb{R}$};

    \node (imgB) [anchor=west, right of=imgA, node distance=\skipimage cm]
    {\includegraphics[
        width=\imagewidth cm,
        trim={1.5cm 1.25cm 1.5cm 1.35cm},  
        clip
    ]{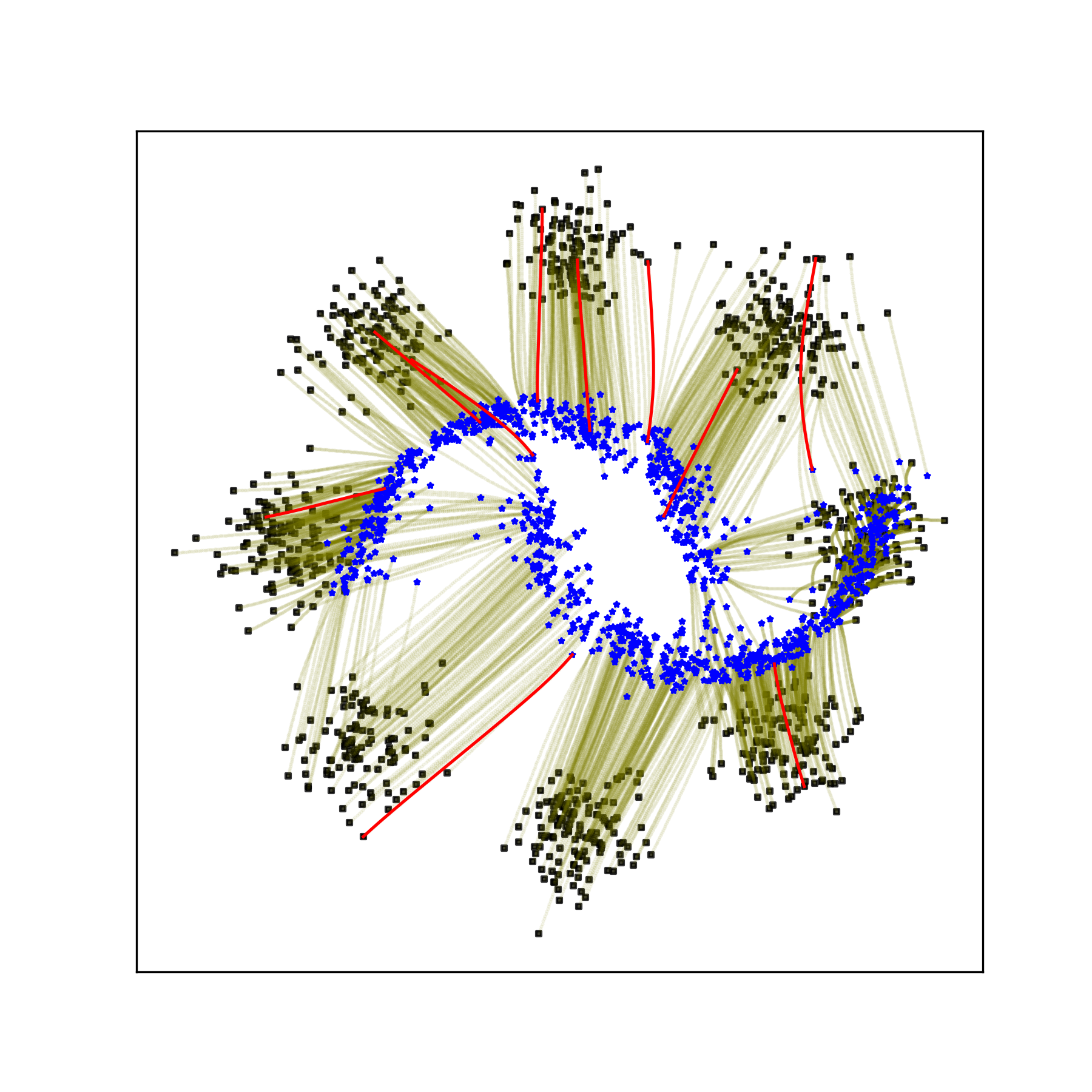}};
    \node[anchor=south, above of=imgB, node distance=\skiptitle cm]
         {OT-Flow Matching};
    \node[anchor=north, below of=imgB, node distance=\skipdesc cm]
         {$v_{\theta}(x, t) \in \mathbb{R}^d$};

    \node (imgC) [right of=imgB, node distance=\skipimage cm]
    {\includegraphics[width=\imagewidth cm]{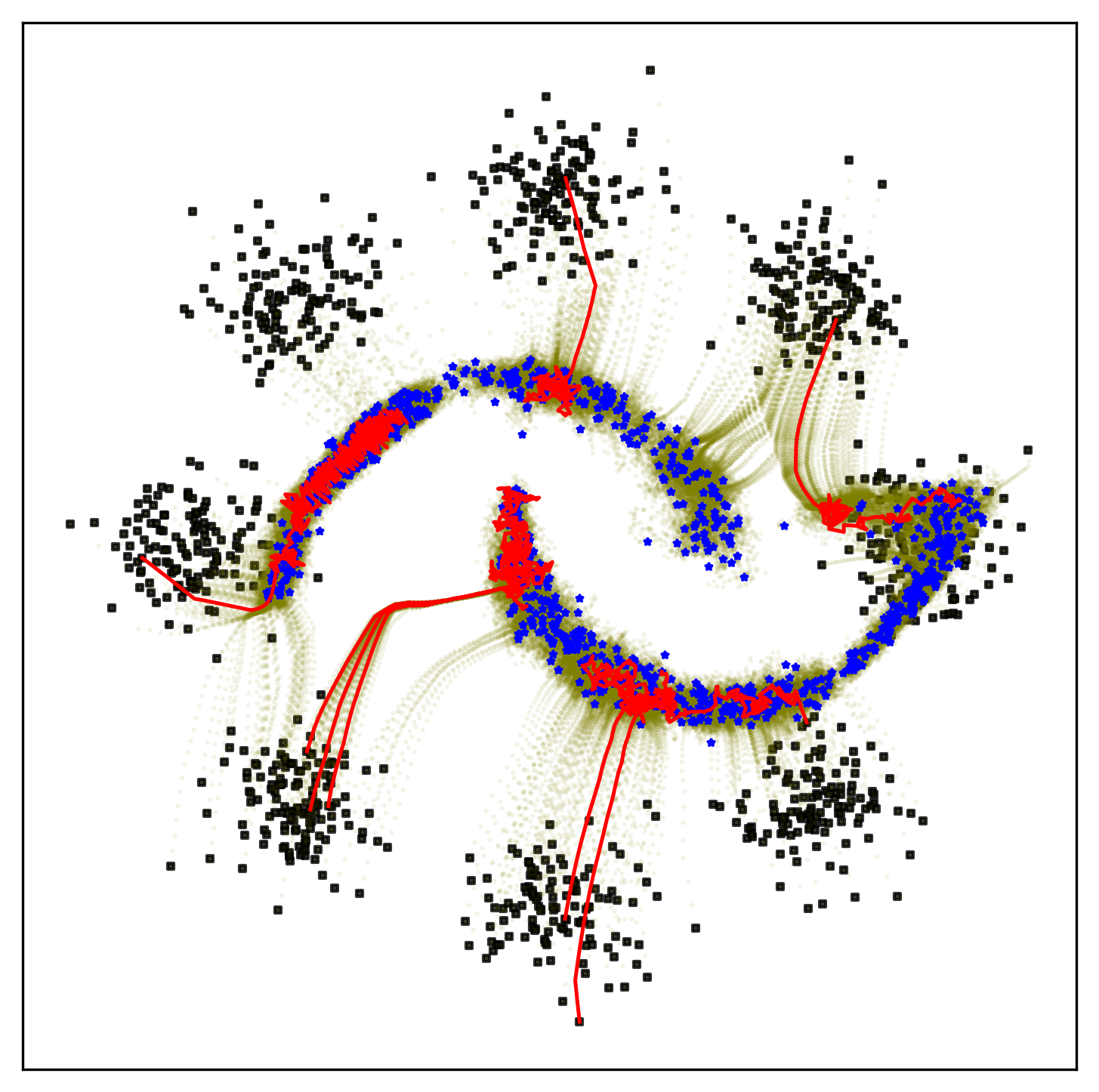}};
    \node[anchor=south, above of=imgC, node distance=\skiptitle cm]
         {Energy-Based Model};
    \node[anchor=north, below of=imgC, node distance=\skipdesc cm]
         {$E_{\theta}(x) \in \mathbb{R}$};

    \node (imgD) [right of=imgC, node distance=\skipimage cm]
    {\includegraphics[width=\imagewidth cm]{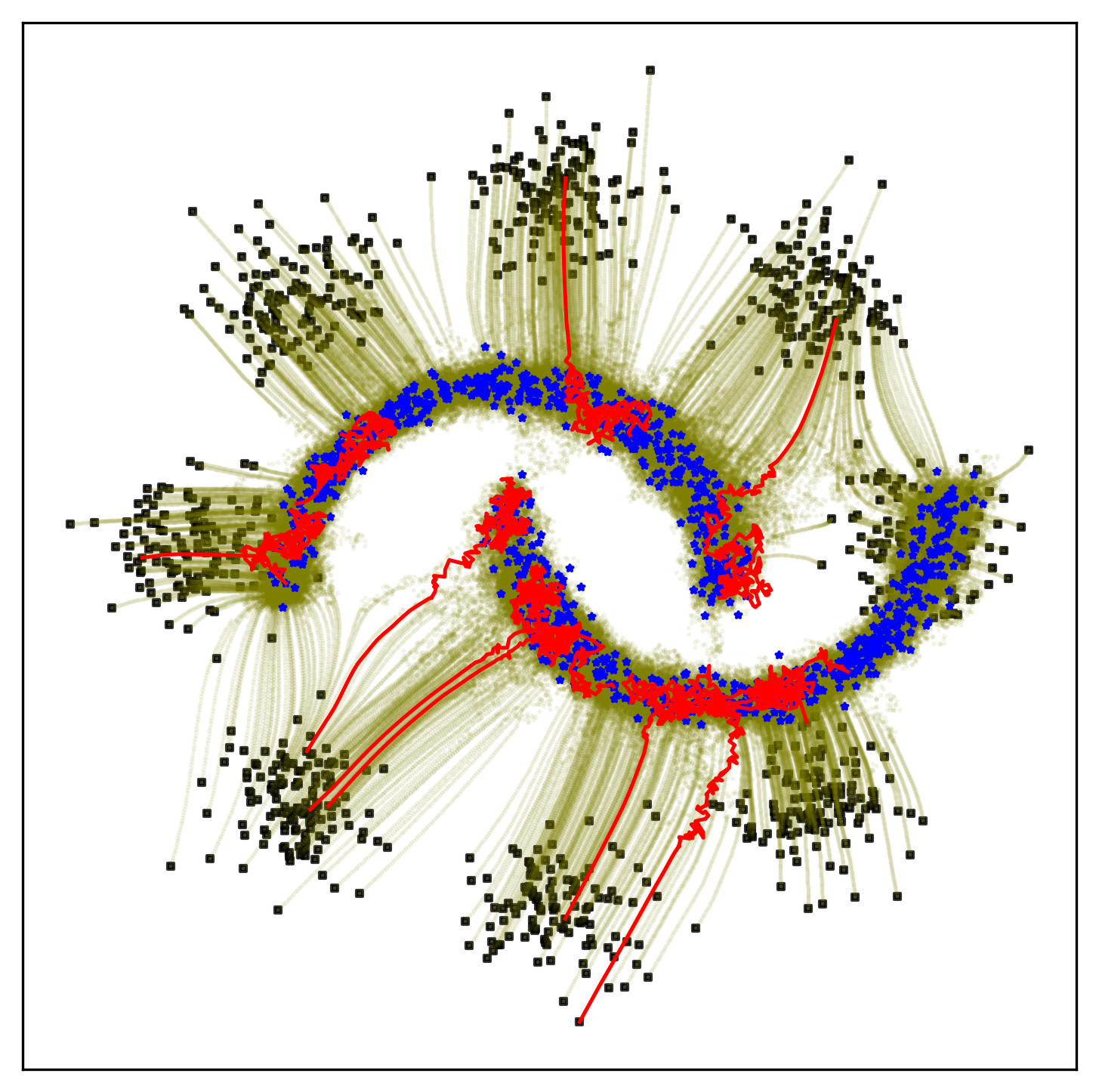}};
    \node[anchor=south, above of=imgD, node distance=\skiptitle cm]
         {Energy Matching  (Ours)};
    \node[anchor=north, below of=imgD, node distance=\skipdesc cm]
         {$V_{\theta}(x) \in \mathbb{R}$};

\end{tikzpicture}

    \caption{Trajectories (green lines) of samples traveling from a noise distribution (black dots; here, a Gaussian mixture model) to a data distribution (blue dots; here, two moons as in \citep{tongimproving}) under four different methods: Action Matching \citep{pmlr-v202-neklyudov23a}, Flow Matching (OT-CFM) \citep{tongimproving}, \glspl*{ebm} trained via contrastive divergence \citep{hinton2002training}, and our proposed Energy Matching. We highlight several individual trajectories in red to illustrate their distinct behaviors. Both Action Matching and Flow Matching learn time-dependent transports and are not trained for traversing the data manifold. Conversely, \glspl*{ebm} and Energy Matching are driven by time-independent fields that can be iterated indefinitely, allowing trajectories to navigate across modes. While samples from \glspl*{ebm} often require additional steps to equilibrate (see, e.g., the visible mode collapses that slow down sampling from the data manifold), Energy Matching directs samples toward the data distribution in ``straight'' paths, without hindering the exploration of the data manifold.}
    \label{fig:cover}
\end{figure}

Generative models learn to map from a simple, easy-to-sample distribution, such as a Gaussian, to a desired data distribution. They do so by
approximating the \gls*{ot} map—such as in flow matching \citep{lipmanflow,liuflow,albergo2023building}—or through iterative noising and denoising schemes, such as in diffusion models
\citep{ho2020denoising, song2021score}.
In addition to being highly effective in sample generation, diffusion- and flow-based models have also been used as priors to regularize poorly posed inverse problems \citep{chung2022diffusion,mardani2023variational,ben2024d}. However, these models do not explicitly capture the unconditional data score and
instead model the score of smoothed manifolds at different noise levels.
The measurement likelihood, on the other hand, is not tractable on these noised manifolds.
As a result, existing approaches repeatedly shuttle between noised and data distributions, 
leading to crude approximations of complex, intractable terms \cite{daras2024survey}. For example, DPS \citep{chung2022diffusion} approximates an intractable integral using a single sample. More recently, D-Flow \citep{ben2024d} optimizes initial noise by differentiating through the simulated trajectory. To the best of our knowledge, these models lack a direct way to navigate the data manifold in search of the optimal solution without repeatedly transitioning between noised and data distributions.

\Glspl*{ebm} \citep{hopfield1982neural,hinton2002training,lecun2006tutorial}
provide an alternative approach for approximating the data distribution by learning a scalar-valued function \(E(x)\)
that specifies an \emph{unnormalized} density \(p(x) \propto \exp\left(-E(x)\right)\).
Rather than explicitly mapping noise samples onto the data manifold, \glspl*{ebm} assign low energies
to regions of high data concentration and high energy elsewhere. This
defines a Boltzmann distribution from which one can sample, for example, via
\emph{Langevin sampling}. In doing so, \glspl*{ebm} explicitly retain the likelihood information in
\(E(x)\). This likelihood information can then be used in conditional generation (e.g., to solve inverse problems), possibly together with additional priors simply by adding their energy terms \citep{du2019implicit}. 
Moreover, direct examination of local curvature on the data manifold—allows the computation of \gls*{lid} (an important proxy for data complexity)—whereas diffusion models can only approximate such curvature in the proximity of noise samples.

Despite the theoretical elegance of using a \emph{single}, time-independent scalar energy, 
practical \glspl*{ebm} have historically suffered from poor generation quality, 
falling short of the performance of diffusion or flow matching models. Traditional methods 
\citep{song2021train} for training \glspl*{ebm}, such as contrastive divergence via \gls*{mcmc} or local score-based approaches \citep{song2019generative}, often 
fail to adequately explore the energy landscape in high-dimensional spaces, leading to 
instabilities and mode collapse. Consequently, many methods resort to time-conditioned 
ensembles \citep{gaolearning}, hierarchical latent ensembles \citep{cui2024learning}, 
or combine \glspl*{ebm} with separate generator networks  trained in cooperation \citep{guo2023egc,zhang2024flow,yoon2024maximum}, 
thereby requiring significantly higher parameter counts and training complexity. 
\clearpage
\begin{mdframed}[hidealllines=true,backgroundcolor=blue!5]
\paragraph{Contributions.}
In this work, we propose \textit{Energy Matching}, a two-regime training strategy that combines the strengths of \glspl*{ebm} and flow matching; see \Cref{fig:cover}. 

When samples lie far from the data manifold, they are efficiently transported toward the data. Once near the data manifold, the flow transitions into Langevin steps governed by an internal energy component, enabling precise exploration of the Boltzmann-like density well around the data distribution. This straightforward approach produces a \emph{time-independent} scalar energy field whose gradient both accelerates sampling and shapes the final density well—via a contrastive objective that directly learns the score at the data manifold—yet remains efficient and stable to train. Empirically, our method significantly outperforms existing \glspl*{ebm} on both CIFAR-10 and ImageNet generation in terms of fidelity, and compares favorably to flow-matching and diffusion models—without auxiliary generators or time-dependent \gls*{ebm} ensembles.

Our proposed process complements the advantages of flow matching with an explicit likelihood modeling, enabling traversal of the data manifold without repeatedly shuffling between noise and data distributions. This simplifies both inverse problem solving and controlled generations under a prior. In addition, to encourage diverse exploration of the data distribution, we showcase how repulsive interaction energies can be easily and effectively incorporated, with an application to conditional protein generation. Finally, we also showcase how analyzing the learned energy reveals insight on the \gls*{lid} of the data with fewer approximations than diffusion models.\footnote{Code repository: \url{https://github.com/m1balcerak/EnergyMatching}}

\end{mdframed}

\section{Energy matching}
In this section, we show how a scalar potential \(V(x)\) can simultaneously provide an optimal-transport-like flow from noise to data while also yielding a Boltzmann distribution that explicitly captures the unnormalized log‑likelihood of the data.

\paragraph{The \texorpdfstring{\gls*{jko}}{Jordan-Kinderlehrer-Otto} scheme.}
The starting point of our approach is the JKO scheme \citep{jordan1998variational}, which is the basis of the success of numerous recent generative models \citep{xu2024normalizing,terpin2024learning,choi2024scalable}. The \gls*{jko} scheme describes the discrete-time evolution of a probability distribution $\rho_t$ along energy-minimizing trajectories in the Wasserstein space,
\begin{equation}
\label{eq:jko}
\rho_{t+\Delta t}
=
\arg\min_{\rho}
\underbrace{\frac{W_2^2\left(\rho,\rho_t\right)}{2\Delta t}}_{\text{Transport Cost}}
+
\underbrace{\int V_{\theta}(x)\mathrm{d}\rho(x)}_{\text{Potential Energy}}
+
\underbrace{\varepsilon(t) \int \rho(x)\log\rho(x)\mathrm{d}x}_{\text{Internal Energy (-\text{Entropy)}}}.
\end{equation}
Here, \(\theta\) denotes the learnable parameters of the scalar potential \(V_{\theta}(x)\), and \(\varepsilon(t)\) is a temperature-like parameter tuning the entropic term. The transport cost is given by the Wasserstein distance,
\begin{equation}
\label{eq:ot}
W_2^2(\rho, \rho_t) = \min_{\gamma \in \Gamma(\rho, \rho_t)}
\int_{\mathbb{R}^d\times\mathbb{R}^d} \norm{x - y}^2 \mathrm{d}\gamma(x, y),
\end{equation}
where $\Gamma(\rho, \rho_t)$ is the set of couplings between $\rho$ and $\rho_t$, i.e., the set of probability distributions on $\mathbb{R}^d\times\mathbb{R}^d$ with marginals $\rho$ and $\rho_t$. Here, $d$ is the dimensionality of the data. Henceforth, we call \gls*{ot} coupling any $\gamma_t$ that yields the minimum in \eqref{eq:ot}. When $\gamma_t = (\mathrm{id}, T)_\#\rho$, i.e., it is the \emph{pushforward} of the map $x \mapsto (x, T(x))$ for some function $T$, we say that $T$ is an \gls*{ot} map from $\rho$ to $\rho_t$.

Differently from most literature, we consider $\varepsilon(t)$ to be dependent on time and study the behavior of \cref{eq:jko} as $t \to \infty$. To fix the ideas, consider, for instance, a linear scheduling:    
\begin{equation} \label{eq:eff_temp}
\varepsilon(t) = 
\begin{cases}
0, & 0 \le t < \Tot, \\
\varepsilon_{\max}\frac{t - \Tot}{1 - \Tot}, & \Tot \le t < 1, \\
\varepsilon_{\max}, & t \ge 1.
\end{cases}
\end{equation}

\paragraph{First-order optimality conditions.}
Following \cite{terpin2024learning}, we analyze \eqref{eq:jko} at each time $t$ via its first-order optimality conditions \citep{lanzetti2024variational,lanzetti2025first}. These conditions characterize the properties of the desired solution and thus represent the optimization goal:

\begin{equation}
\label{eq:first-order}
    \frac{1}{\Delta t}(x - y)
    +
    \nabla_x V_{\theta}(x)
    +
    \varepsilon(t) \nabla_x \log \rho_{t+\Delta t}(x)
    =0,\qquad(x, y) \in \mathrm{supp}(\gamma_t)
\end{equation}
where \(\gamma_t\) is an \gls*{ot} plan between the distributions $\rho_{t+\Delta t}$ and $\rho_{t}$ and $\mathrm{supp}(\gamma_t)$ is the support of $\gamma_t$. That is, this condition has to hold for all pairs of points in the support of $\rho_{t+\Delta t}$ and $\rho_t$ that are coupled by \gls*{ot}.
Intuitively, analyzing \eqref{eq:first-order} provides us with two key insights:
\begin{enumerate}[leftmargin=*]
    \item For times $t < \Tot$, $\varepsilon(t) = 0$ and \eqref{eq:first-order} becomes
    \begin{equation}
    \label{eq:onlypotential}
    \frac{1}{\Delta t}(x - y)
    +
    \nabla_x V_{\theta}(x)
    =0\qquad(x, y) \in \mathrm{supp}(\gamma_t).
    \end{equation}
    That is, the system is in an \gls*{ot}, flow-like, regime.
    \item Near the data manifold, which we aim at modeling with the equilibrium distribution $\rho_\text{eq}$ of \eqref{eq:jko}, $\rho_{t+\Delta t} \approx \rho_\text{eq}$ and, thus, for $t \gg 1$, $x \approx y$ for all $(x, y) \in \mathrm{supp}(\gamma_t)$. Then, we can simplify \eqref{eq:first-order} as
    \[
    \varepsilon_{\max} \nabla_x \log \rho_{\mathrm{eq}}(x)
    =
    -\nabla_x V_{\theta}(x)
    \quad\Longrightarrow\quad
    \rho_{\mathrm{eq}}(x)
    \propto
    \exp\left(-\tfrac{V_{\theta}(x)}{\varepsilon_{\max}}\right).
    \]
    Thus, the equilibrium distribution is described by an \gls*{ebm}, \(\exp\left(-E(x)\right)\), with \(E(x)=\frac{V_{\theta}(x)}{\varepsilon_{\max}}\).
\end{enumerate}

\paragraph{Our approach in a nutshell.}
Combining the two insights above, we propose a generative framework that combines \gls*{ot} and \glspl*{ebm} to learn a \emph{time-independent} scalar potential \(V_{\theta}(x)\) whose Boltzmann distribution,
\begin{equation}
    \rho_{\mathrm{eq}}(x) \propto \exp\left(-\frac{V_{\theta}(x)}{\varepsilon_{\max}}\right),
\end{equation}
matches \(\rho_{\mathrm{data}}\). To transport samples efficiently from noise \(\rho_0\) to \(\rho_{\mathrm{eq}}\approx \rho_{\mathrm{data}}\), we use two regimes:
\begin{itemize}[leftmargin=*]
    \item \textit{Away from the data manifold}: \(\varepsilon \approx 0\). The flow is deterministic and OT-like, allowing rapid movement across large distances in sample space.
    \item \textit{Near the data manifold}: \(\varepsilon \approx \varepsilon_{\max}\). Samples diffuse into a stable Boltzmann distribution, properly covering all data modes.
\end{itemize}
By combining the long-range transport capability of flows with the local density modeling flexibility of \glspl*{ebm}, we achieve tractable sampling and explicitly encode the unnormalized log-likelihood $-V_\theta(x)/\varepsilon_{\max}$ of the underlying data distribution; see \cref{fig:cover}.

\subsection{Training objectives}
In practice, we balance the two objectives by initially training $V_{\theta}$ exclusively with the optimal-transport-like objective ($\varepsilon = 0$, see \cref{sec:pre-training}), ensuring a stable and consistent generation of high-quality negative samples for the contrastive phase. Subsequently, we jointly optimize both the transport-based and contrastive divergence objectives, progressively increasing the effective temperature to $\varepsilon = \varepsilon_{\max}$ as samples approach the data manifold (i.e., the equilibrium distribution); see \cref{sec:fine-tuning}.

\subsubsection{Flow-like objective $\Lot$}
\label{sec:pre-training}
We begin by constructing a global velocity field \(-\nabla_x V_{\theta}(x)\) that carries noise samples \(\{x_0\}\) to data samples \(\{x_{\mathrm{data}}\}\) with minimal detours. For this, we consider geodesics in the Wasserstein space \citep{Ambrosio2008GradientMeasures}. Practically, we compute the \gls*{ot} coupling \(\gamma^*\) between two uniform empirical probability distributions, one supported on a mini-batch of the data, and one supported on a set of noise samples with the same cardinality. These samples are drawn from an easy-to-sample distribution; in our case, a Gaussian. Since the probability distributions are uniform and empirical with the same number of samples, a transport map $T$ is guaranteed to exist \citep{Ambrosio2008GradientMeasures}.

\begin{remark}[\gls*{ot} solver]
Depending on the method used to compute the \gls*{ot} coupling, an explicit \gls*{ot} map may or may not be obtained. Similarly, if the number of noise samples differs from the mini-batch size $B$, the resulting \gls*{ot} coupling generally will not correspond to a map. In this case, one can adapt the algorithm by defining a threshold $\pi_\mathrm{th}$ and considering all pairs $(x_\mathrm{data}, x_0)$ for which the coupling value satisfies $\gamma^\ast(x_\mathrm{data}, x_0) > \pi_\mathrm{th}$. 
In our experiments, we used the \texttt{POT} solver~\citep{Flamary2021} and did not observe benefits from using a sample size different from $B$, consistent with previous approaches~\citep{tongimproving}.
\end{remark}

Then, for each data point $x_\mathrm{data}$ we define the \emph{interpolation}
$
x_t = (1 - t) T(x_{\mathrm{data}}) + t x_\mathrm{data}
$,
which is a point along the geodesic. The \emph{velocity} of each $x_t$ is $x_{\mathrm{data}} - T(x_{\mathrm{data}})$ (i.e., the samples move from the noise to the data distribution at constant speed) and, in this regime, we would like to have $-\nabla_x V_\theta(x_t) \approx x_{\mathrm{data}} - T(x_{\mathrm{data}})$. For this, we define the loss:
\[
\Lot
=
\mathbb{E}{\substack{x_\mathrm{data} \in \mathcal{D}\\t \sim U(0,\Tot)}}\left[\norm{\nabla_x V_{\theta}(x_{t}) + x_{\mathrm{data}} - T({x_\mathrm{data}})}^2\right].
\]
This objective can be interpreted as a flow-matching formulation under the assumption that the velocity field is both time-independent and given by the gradient of a scalar potential, thereby imposing an irrotational condition. This aligns naturally with \gls*{ot}, which also yields an irrotational velocity field—any rotational component would add unnecessary distance to the flow and thus inflate the transport cost without benefit. Our experimental evidence adds to the recent study by \citep{sun2025noise}, in which the authors observed that time-independent velocity fields can, under certain conditions, outperform time-dependent noise-conditioned fields in sample generation. 

\begin{algorithm}
\small
\caption{Phase 1 (warm-up).}
\label{alg:pretrain_revised}
\begin{algorithmic}[1]
\State \textbf{Initialize} model parameters \(\theta\)
\For{iteration \(n = 0, 1, \ldots\)}
  \State Sample mini-batch \(\{x_{\mathrm{data}, b}\}_{b=1}^B \sim \mathcal{D}\)
  \Comment{Data samples}
  \State Sample mini-batch \(\{x_{0,b}\}_{b=1}^B \sim \mathcal{N}(0, I)\)
  \Comment{Random Gaussian samples}
  \State \(T \gets \mathrm{OTsolver}(\{x_{\mathrm{data}, b}\}, \{x_{0,b}\} )\)
  \Comment{Compute OT map}
  \State Sample \(\{t_b\}_{b=1}^B \sim U(0,\Tot)\) \Comment{Typically $\Tot = 1$ for the warm-up}
  \State  Set interpolations \( x_{t_b} \gets (1 - t_b)\,T(x_{\mathrm{data},b}) + t_b\,x_{\mathrm{data},b} \)
  \Comment{Interpolation along geodesics}
  \State \(\Lot(\theta) \gets \sum_{b=1}^{B} \|\nabla_x V_\theta(x_{t_b}) + x_{\mathrm{data}, b} - T(x_{\mathrm{data}, b})\|^2\)
  \Comment{Loss function}
  \State \(\theta \gets \theta - \alpha \nabla_{\theta} \Lot(\theta)\)
  \Comment{Gradient update with learning rate $\alpha$}
\EndFor
\State \Return \(\theta\)\Comment{Trained $\theta$}
\end{algorithmic}
\end{algorithm}

\subsubsection{Contrastive objective\texorpdfstring{ $\Lcd$}{}}
\label{sec:fine-tuning}
Near the data manifold, \(V_{\theta}(x)\) is refined so that
\(\rho_{\mathrm{eq}}(x) \propto \exp\left(-V_{\theta}(x)/\varepsilon_{\max}\right)\)
matches the data distribution. We adopt the contrastive divergence loss described in \glspl*{ebm}~\citep{hinton2002training},
\[
\Lcd
=
\mathbb{E}_{x \sim p_{\text{data}}}\left[\frac{V_{\theta}(x)}{\varepsilon_{\max}}\right]
-
\mathbb{E}_{\tilde{x} \sim \mathrm{sg}(p_{\mathrm{eq}})}\left[\frac{V_{\theta}(\tilde{x})}{\varepsilon_{\max}}\right],
\]
where \(\tilde{x}\) are ``negative'' samples of the equilibrium distribution induced by \(V_\theta\). We approximate these samples using an \gls*{mcmc} Langevin chain~\citep{welling2011bayesian}. We split the initialization for negative samples: half begin at real data, and half begin at the noise distribution. This way, \(V_{\theta}(x)\) forms well-defined basins around high-density regions while also shaping regions away from the manifold, correcting the generation. The \(\mathrm{sg}(\cdot)\) indicates a \emph{stop-gradient} operator, which ensures gradients do not back-propagate through the sampling procedure.

\subsubsection{Dual objective and implementation notes}

To balance the deterministic flow-like regime (where $\varepsilon \approx 0$) away from the data manifold and the stochastic Boltzmann regime (where $\varepsilon \approx \varepsilon_{\max}$) near equilibrium, we adopt the linear temperature schedule described in \eqref{eq:eff_temp}.
We introduce a dataset-specific hyperparameter $\lambda_{\mathrm{CD}}$ to stabilize the contrastive objective by appropriately weighting $\Lcd$ relative to $\Lot$.
The resulting algorithm is described in detail in \cref{alg:pretrain_revised} and \cref{alg:fine_tune_revised}.  
Since \cref{alg:fine_tune_revised} benefits from high-quality negatives, we begin with \cref{alg:pretrain_revised} (and, thus, with $\Lot$ only) to ensure sufficient mixing of noise-initialized negatives.  

Given the trained models, we define a sampling time $\samplingTime$. Although convergence to the equilibrium distribution is guaranteed only as \(\samplingTime \to \infty\), we empirically observe that sample quality (measured with \gls*{fid}) plateaus by \(\samplingTime = 3.25\) on CIFAR-10; see \cref{appendix:ablations}. The sampling procedure, which optionally includes conditional and interaction terms, is detailed in \cref{alg:sampling_inverse_interaction}. In practice, we implement training using explicit Euler--Maruyama updates and sampling with an Euler--Heun predictor-corrector scheme, while for simplicity the algorithms illustrate only explicit updates. Additionally, the constant factor \(1/\varepsilon_{\max}\) in \(\Lcd\) is absorbed into \(\lambda_{\mathrm{CD}}\).

\cref{appendix:energy-landscape} discusses how the landscape of \(V_{\theta}\) evolves across these two phases. Hyperparameters for each dataset, along with intuitions to guide their selection for new datasets, are provided in \cref{appendix:training-details}.

\begin{algorithm}
\small
\caption{Phase 2 (main training).}
\label{alg:fine_tune_revised}
\begin{algorithmic}[1]
\State \(\theta \gets \theta_{\text{pretrained}}\) \Comment{Initialize from Algorithm~\ref{alg:pretrain_revised}}

\For{iteration \(n = 0, 1, \ldots\)}
  \State \(\Lot \leftarrow\) Use lines 3--8 from Algorithm~\ref{alg:pretrain_revised}

  \State Initialize negative samples \(\{x_{\mathrm{neg}, b}^{(0)}\}_{b=1}^B\) from noise and/or data
  \Comment{Negative samples}

  \For{\(m = 0, 1, \ldots, M_{\text{Langevin}}-1\)}
    \For{\(b = 1\) to \(B\)}
        \State \(\varepsilon^{(m)} \gets 
        \begin{cases}
            \varepsilon_{\max}, & \text{if initialized from data (Optional)}\\[2pt]
            \varepsilon(m\Delta t)\text{ from \eqref{eq:eff_temp}}, & \text{otherwise}
        \end{cases}\)
        \State Sample \(\eta_b \sim \mathcal{N}(0, I)\)
        \State \(x_{\mathrm{neg}, b}^{(m+1)} \gets x_{\mathrm{neg}, b}^{(m)} 
           - \Delta t \nabla_x V_{\mathrm{sg}(\theta)}(x_{\mathrm{neg}, b}^{(m)}) 
           + \sqrt{2 \Delta t \varepsilon^{(m)}}\,\eta_b\)
        \Comment{Langevin dynamics step}
    \EndFor
  \EndFor
  
  \State \(\Lcd \gets \frac{1}{B}\sum_{b=1}^B \left[V_{\theta}(x_{\mathrm{data}, b}) - V_{\theta}(x_{\mathrm{neg}, b}^{(M_{\text{Langevin}})})\right]\)
  \Comment{Contrastive divergence loss}

  \State \(\mathcal{L}(\theta) \gets \Lot + \lambda_{\mathrm{CD}} \Lcd\)

  \State Update \(\theta \gets \theta - \alpha \nabla_\theta \mathcal{L}(\theta)\)
  \Comment{Gradient descent step}
\EndFor

\State \Return \(\theta\) \Comment{Trained $\theta$}
\end{algorithmic}
\end{algorithm}

\section{Applications}
\label{sec:results}
In this section, we demonstrate the effectiveness and versatility of our proposed Energy Matching approach across three applications: (i) unconditional generation (ii) inverse problems, and (iii) \gls*{lid} estimation. The model architecture and all the training details are reported in \cref{appendix:training-details}.
\subsection{Unconditional generation}
\label{sec:results:unconditional}
We compare four classes of generative models: (1) {Diffusion models}, which deliver state-of-the-art quality but typically require many sampling steps; (2) {Flow-based methods}, which learn \gls*{ot} paths for more efficient sampling with fewer steps; (3) {\glspl*{ebm}}, which directly model the log-density as a scalar field, offering flexibility for inverse problems and constraints but sometimes at the expense of sample quality; and (4) {Ensembles (Diffusion with one or many \glspl*{ebm})}, which combine diffusion’s robust sampling with elements of \gls*{ebm} flexibility but can become large and complex to train.
Our approach, {Energy Matching}, offers a simple (a single time-independent scalar field) yet powerful \gls*{ebm}-based framework.
We evaluate our approach on CIFAR-10 \citep{krizhevsky2009learning} and ImageNet32x32 \citep{deng2009imagenet,chrabaszcz2017downsampled} datasets, reporting \gls*{fid} scores in \cref{tab:fid_comparison} and \cref{tab:fid_comparison_imagenet}, respectively. Our method outperforms state-of-the-art \glspl*{ebm}, reducing the \gls*{fid} score by more than $50\%$.

\begin{table}
\centering
\caption{\small FID$\downarrow$ score comparison for unconditional CIFAR-10 generation (lower is better). Unless otherwise specified, we use results for solvers that most closely match our setup (325 fixed-step Euler–Heun \citep{butcher2016}). $^*$ indicates reproduced methods, while all other entries reflect the best reported results. EGC in its unconditional version has been reported in \citep{zhulearning}}.
\label{tab:fid_comparison}
\scriptsize
\begin{tabular}{l C | l C}
\toprule
\rowcolor{gray!35}
\multicolumn{2}{c|}{\textbf{Learning Unnormalized Data Likelihood}} & \multicolumn{2}{c}{\textbf{Learning Transport/Score Along Noised Trajectories}} \\
\midrule
\rowcolor{gray!10}
\multicolumn{2}{c|}{\textbf{Ensembles: Diffusion + (one or many) EBMs}} & \multicolumn{2}{c}{\textbf{Diffusion Models}} \\
\midrule
Hierarchical EBM Diffusion \citep{cui2024learning} & 8.93 & DDPM$^*$ \citep{ho2020denoising} & 6.45 \\
EGC \citep{guo2023egc} & 5.36 & DDPM++ \textit{(62M params, 1000 steps)} \citep{kim2021soft} & 3.45 \\
Cooperative DRL \textit{(40M params)} \citep{zhulearning} & 4.31 &  NCSN++ \textit{(107M params, 1000 steps)} \citep{song2021score} & 2.45   \\
Cooperative DRL-large \textit{(145M params)} \citep{zhulearning} & 3.68 &   &    \\
\midrule
\rowcolor{gray!10}
\multicolumn{2}{c|}{\textbf{Energy-based Models}} & \multicolumn{2}{c}{\textbf{Flow-based Models}} \\
\midrule
ImprovedCD \citep{du2021improved} & 25.1 & Action Matching \citep{pmlr-v202-neklyudov23a} & 10.07 \\
CLEL-large \textit{(32M params)} \citep{lee2023guiding} & 8.61 & Flow-matching \citep{lipmanflow} & 6.35 \\
\cellcolor{blue!10}Energy Matching \textit{(50M params, \textbf{Ours})} & \cellcolor{blue!10}3.34 & OT-CFM$^*$ \textit{(37M params)} \citep{tongimproving} & 4.04 \\
\bottomrule
\end{tabular}
\end{table}

\begin{table}
\centering
\caption{\small FID$\downarrow$ score comparison for unconditional ImageNet 32x32 generation (lower is better). Unless otherwise specified, we use results for solvers that most closely match our setup (300 fixed-step Euler–Heun \citep{butcher2016}).}
\label{tab:fid_comparison_imagenet}
\scriptsize
\begin{tabular}{l C | l C}
\toprule
\rowcolor{gray!35}
\multicolumn{2}{c|}{\textbf{Learning Unnormalized Data Likelihood}} & \multicolumn{2}{c}{\textbf{Learning Transport/Score Along Noised Trajectories}} \\
\midrule
\rowcolor{gray!10}
\multicolumn{2}{c|}{\textbf{Ensembles: Diffusion + (one or many) EBMs}} & \multicolumn{2}{c}{\textbf{Diffusion Models}} \\
\midrule
Cooperative DRL \textit{(40M params)}  \citep{zhulearning} & 9.35 & DDPM++ \textit{(62M params, 1000 steps)} \citep{kim2021soft} & 8.42 \\
\midrule
\rowcolor{gray!10}
\multicolumn{2}{c|}{\textbf{Energy-based Models}} & \multicolumn{2}{c}{\textbf{Flow-based Models}} \\
\midrule
ImprovedCD \citep{du2021improved}  & 32.48 & Flow-matching \citep{lipmanflow} \textit{(196M params)}& 5.02 \\
CLEL-base \citep{lee2023guiding} \textit{(7M params)} & 22.16 & &  \\
CLEL-large \citep{lee2023guiding} \textit{(32M params)} & 15.47 &  &  \\
\cellcolor{blue!10}Energy Matching \textit{(50M params, \textbf{Ours})} & \cellcolor{blue!10} 6.64 &  &  \\
\bottomrule
\end{tabular}
\end{table}
\subsection{Inverse problems}
\label{sec:results:conditional}
In many practical applications, we are interested in recovering some data \(x\) from noisy measurements \(y\) generated by an operator \(A\), $y=A(x)+w$, where $w \sim \mathcal N\!\bigl(0,\sqrt{2}\,\zeta I\bigr)$. In this setting, the {posterior} distribution of $x$ given $y$ is
\begin{equation}
\label{eq:inverse-problem-prob}
p(x | y) \propto
\underbrace{\exp\left(-\frac{1}{\zeta^2}\|y - A(x)\|^2\right)}_{\displaystyle \propto p(y | x)}
\underbrace{\exp\left(-E_\theta(x)\right)}_{\displaystyle \propto p(x)},
\end{equation}

where $E_\theta(x)$ is an energy function which one can learn from the data, an \gls*{ebm}. Because we want to {sample} $x$ given a measurement $y$, this reconstruction task is often referred to as an \textit{inverse problem}. 
Here, $\|y - Ax\|^2$ encodes the measurement fidelity with $\zeta$  controlling the balance between this fidelity term and the prior. We obtain the prior term $E_{\theta}(x) = \frac{V_\theta(x)}{\varepsilon_{\max}}$ by training $V_\theta(x)$ via Energy Matching.
Samples from this posterior can be drawn by starting from a random sample $x^{(0)} \sim \mathcal{N}(0, I)$ and following a Langevin update. We detail the algorithm for generating solutions to inverse problems in \cref{alg:sampling_inverse_interaction} (which also incorporates additional interaction energy $W(x,x')$ between generated samples). We demonstrate our model's capabilities qualitatively through a controlled inpainting task and quantitatively via a protein inverse design benchmark. Specific hyperparameters are detailed in \cref{appendix:training-details}.

\begin{algorithm}[H]
    \small
\caption{Unconditional/conditional sampling with optional interaction energy}

    \label{alg:sampling_inverse_interaction}
    \begin{algorithmic}[1]
        \For{\(m = 1\) to \(M\)}
            \State Initialize \(x_m^{(0)}\) from noise and/or data
                \Comment{Initialize each chain}
        \EndFor
        
        \State \(N \gets \lfloor \samplingTime / \Delta t \rfloor\) 
            \Comment{Number of Langevin steps for sampling time \(\samplingTime\)}

        \For{\(n = 0, 1, \ldots, N - 1\)}
            \For{\(m = 1, 2, \ldots, M\)}
                \Comment{Prior + data fidelity + interaction}

                \State \(\varepsilon^{(n)} \gets 
                    \begin{cases}
                        \varepsilon_{\max}, & \text{if initialized from data (Optional)}\\[2pt]
                        \varepsilon(n\Delta t)\text{ from \eqref{eq:eff_temp}}, & \text{otherwise}
                    \end{cases}\)

                \State \(U_\theta\bigl(x_m^{(n)}\bigr) \gets
                    V_\theta\bigl(x_m^{(n)}\bigr) 
                    + \varepsilon^{(n)}\bigl\lVert y - A\bigl(x_m^{(n)}\bigr)\bigr\rVert^2 / \zeta^2
                    + \varepsilon^{(n)}\sum_{k \neq m}W\bigl(x_m^{(n)}, x_k^{(n)}\bigr)\)
                    
                \State Sample \(\eta_m^{(n)} \sim \mathcal{N}(0, I)\)
                \State \(x_m^{(n+1)} \gets x_m^{(n)} 
                    - \Delta t\,\nabla_x U_\theta\bigl(x_m^{(n)}\bigr) 
                    + \sqrt{2\varepsilon^{(n)}\Delta t}\;\eta_m^{(n)}\) 
                    \Comment{Langevin dynamics step}
            \EndFor
        \EndFor
        
        \State \Return \(\{x_m^{(N)}\}_{m=1}^M\) 
            \Comment{Final samples}
    \end{algorithmic}
\end{algorithm}

\begin{figure}[h!]
    \centering
    \begin{tikzpicture}[node distance=1.3cm]
        \node (corrupted) {\includegraphics[width=2.3cm]{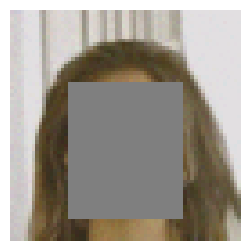}};
        
        \node[right of=corrupted, anchor=south west, node distance=5cm] (reconstructedwithout1) 
        {\includegraphics[width=2.3cm]{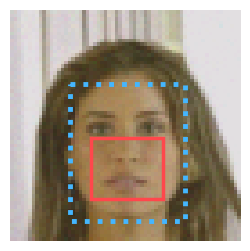}};
        \node[right of=reconstructedwithout1, anchor=west] (reconstructedwithout2) 
        {\includegraphics[width=2.3cm]{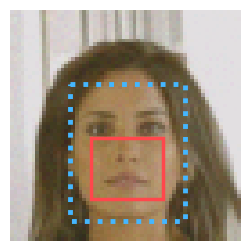}};

        \draw[->, ultra thick] (corrupted) -- 
            node[yshift=.5cm, anchor=south, above, align=center] {\footnotesize
                $\begin{aligned}
                &p(x_1, x_2 | y) \\[-3pt]
                &\propto p(x_1| y)p(x_2| y)
                \end{aligned}$}
            (reconstructedwithout1);

        \node[right of=corrupted, anchor=north west, node distance=5cm] (reconstructedwith1) 
        {\includegraphics[width=2.3cm]{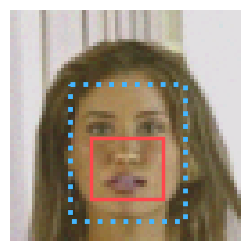}};
        \node[right of=reconstructedwith1, anchor=west] (reconstructedwith2) 
        {\includegraphics[width=2.3cm]{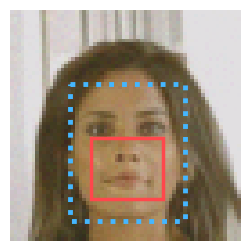}};

        \draw[->, ultra thick] (corrupted) -- 
            node[yshift=-0.5cm, anchor=north, below, align=center] {\footnotesize
                $\begin{aligned}
                    &p(x_1, x_2 | y) \\[-3pt]
                    &\propto p(x_1| y)p(x_2 | y) \\
                    &\quad\cdot\exp\left(\frac{\norm{B\,(x_1 - x_2)}^2}{\sigma^2}\right)
                \end{aligned}$}
            (reconstructedwith1);

    \end{tikzpicture}
    \caption{Controlled inpainting for diverse reconstructions. 
      On the left is the masked face. On the right are two reconstructions:
      the top pair without the interaction term and the bottom pair with it.
      The interaction term applies in the \textcolor{lred}{solid red square}
      (where \(B\) has ones), and the measurement matrix \(A\) is the 
      \textcolor{lblue}{dotted blue square} (zeros inside, ones outside).
      By encouraging \(x_1\) and \(x_2\) to differ in the target region,
      the interaction yields a wider range of completions while preserving 
      fidelity.}
    \label{fig:controlled_inpainting}
\end{figure}

\paragraph{Controlled inpainting.} Suppose we want to recover two images from a masked image while encouraging diverse reconstructions. \glspl*{ebm} allow this by introducing an additional interaction energy, \(W(x_1, x_2) = -\frac{\|B(x_1 - x_2)\|^2}{\sigma^2}\), where \(B\) has ones in the region of interest (focusing diversity there) and zeros elsewhere, and \(\sigma\) is a hyperparameter controlling the interaction's strength. Specifically, we define \(p(x_1, x_2 \mid y) \propto p(x_1 \mid y)\,p(x_2 \mid y)\,\exp\bigl(-W(x_1,x_2)\bigr)\), which gives high probability to pairs \((x_1, x_2)\) that lie far apart in the specified region \(B\). This encourages exploring the edges of the posterior rather than just its modes, and with suitable \(W\), samples shift toward rare events without needing many draws.
To illustrate the interaction term’s advantages for diverse reconstruction, we apply our method to a CelebA \citep{liu2015faceattributes} \(64\times64\) inpainting task. 
As shown in Figure~\ref{fig:controlled_inpainting}, we start from a partially observed (masked) face and aim to reconstruct two distinct high-fidelity completions.

\FloatBarrier

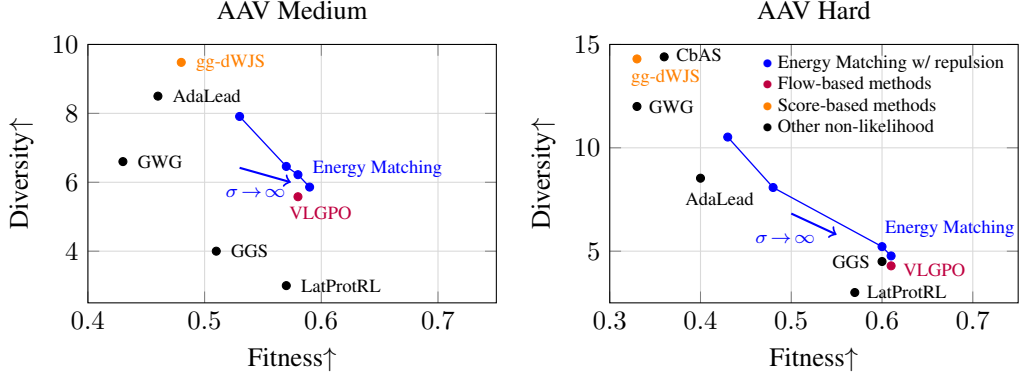
\begin{figure}[htbp]
  \centering
  \begin{tikzpicture}
    \begin{groupplot}[
        group style = {
          group name = aav,
          group size = 2 by 1,
          horizontal sep = 1.5cm
        },
        width=0.5\textwidth,
        height=5cm,
        grid=both,
        major grid style={gray!30},
        minor grid style={gray!10},
      ]

      \nextgroupplot[
        title={AAV Medium},
        xlabel={Fitness$\uparrow$},
        ylabel={Diversity$\uparrow$},
        xmin=0.4, xmax=0.75,
        ymin=2.5, ymax=10.0,
      ]
        \addplot[only marks,mark=*,mark size=1.5pt,draw=blue,fill=blue]
          coordinates {
            (0.59,5.86) (0.58,6.22) (0.57,6.46) (0.53,7.91)
          };
        \addplot[blue,line width=0.5pt,mark=none]
          coordinates {
            (0.59,5.86) (0.58,6.22) (0.57,6.46) (0.53,7.91) 
          };
        \draw[blue,thick,->] plot[smooth] coordinates {
          (axis cs:0.53,6.42) (axis cs:0.574,6.0) 
        };
        \node[blue,font=\scriptsize,anchor=west]
          at (axis cs:0.51,5.70) {$\sigma\!\to\!\infty$};

        \addplot[only marks,mark=*,mark size=1.5pt,draw=purple,fill=purple]
          coordinates {(0.58,5.58)};
        \addplot[only marks,mark=*,mark size=1.5pt,draw=orange,fill=orange]
          coordinates {(0.48,9.48)};
        \addplot[only marks,mark=*,mark size=1.5pt,draw=black,fill=black]
          coordinates {
            (0.57,3.00) (0.51,4.00) (0.46,8.50) (0.43,6.60)
          };

        \node[anchor=west,inner sep=1pt,font=\scriptsize,text=blue]
          at (axis cs:0.59,6.4) {Energy Matching};
        \node[anchor=west,inner sep=1pt,font=\scriptsize,text=purple]
          at (axis cs:0.57,5.10) {VLGPO};
        \node[anchor=west,inner sep=1pt,font=\scriptsize]
          at (axis cs:0.58,3.00) {LatProtRL};
        \node[anchor=west,inner sep=1pt,font=\scriptsize]
          at (axis cs:0.52,4.00) {GGS};
        \node[anchor=west,inner sep=1pt,font=\scriptsize,text=orange]
          at (axis cs:0.49,9.48) {gg-dWJS};
        \node[anchor=west,inner sep=1pt,font=\scriptsize]
          at (axis cs:0.47,8.50) {AdaLead};
        \node[anchor=west,inner sep=1pt,font=\scriptsize]
          at (axis cs:0.44,6.60) {GWG};

      \nextgroupplot[
        title={AAV Hard},
        xlabel={Fitness$\uparrow$},
        ylabel={Diversity$\uparrow$},
        xmin=0.3, xmax=0.75,
        ymin=2.5, ymax=15.0,
      ]
        \addplot[only marks,mark=*,mark size=1.5pt,draw=blue,fill=blue]
          coordinates {
            (0.61,4.77) (0.60,5.22)  (0.48,8.08)
            (0.43,10.52)
          };
        \addplot[blue,line width=0.5pt,mark=none]
          coordinates {
            (0.61,4.77) (0.60,5.22)  (0.48,8.08) (0.43,10.52)
          };
        \draw[blue,thick,->] plot[smooth] coordinates {
          (axis cs:0.50,6.82) (axis cs:0.55,5.8)
        };

        \addplot[only marks,mark=*,mark size=1.5pt,draw=purple,fill=purple]
          coordinates {(0.61,4.29)};
        \addplot[only marks,mark=*,mark size=1.5pt,draw=orange,fill=orange]
          coordinates {(0.33,14.3)};
        \addplot[only marks,mark=*,mark size=1.5pt,draw=black,fill=black]
          coordinates {
            (0.57,3.00) (0.60,4.50) (0.40,8.53)
            (0.36,14.4) (0.33,12.0) (0.10,11.6)
          };
        \node[blue,font=\scriptsize,anchor=west]
          at (axis cs:0.45,5.6) {$\sigma\!\to\!\infty$};

        \node[anchor=west,inner sep=1pt,font=\scriptsize,text=blue]
          at (axis cs:0.60,6.1)   {Energy Matching};
        \node[anchor=west,inner sep=1pt,font=\scriptsize,text=purple]
          at (axis cs:0.62,4.1)    {VLGPO};
        \node[anchor=west,inner sep=1pt,font=\scriptsize]
          at (axis cs:0.58,3.00)   {LatProtRL};
        \node[anchor=east,inner sep=1pt,font=\scriptsize]
          at (axis cs:0.59,4.50)   {GGS};
        \node[anchor=west,inner sep=1pt,font=\scriptsize,text=orange]
          at (axis cs:0.32,13.4)   {gg-dWJS};
        \node[anchor=west,inner sep=1pt,font=\scriptsize]
          at (axis cs:0.38,7.53)   {AdaLead};
        \node[anchor=west,inner sep=1pt,font=\scriptsize]
          at (axis cs:0.37,14.5)   {CbAS};
        \node[anchor=west,inner sep=1pt,font=\scriptsize]
          at (axis cs:0.34,12.0)   {GWG};
        \node[anchor=west,inner sep=1pt,font=\scriptsize]
          at (axis cs:0.10,11.6)   {GFN-AL};

        \node[anchor=north east,inner sep=2pt,align=left,font=\scriptsize]
          at (rel axis cs:0.98,0.98) {%
            \textcolor{blue}{$\bullet$} Energy Matching w/ repulsion\\
            \textcolor{purple}{$\bullet$} Flow-based methods\\
            \textcolor{orange}{$\bullet$} Score-based methods\\
            \textcolor{black}{$\bullet$} Other non-likelihood
          };

    \end{groupplot}
  \end{tikzpicture}
\caption{Fitness–diversity trade-off for protein inverse design on the AAV Medium (left) and Hard (right) benchmarks. 
We compare our Energy Matching method (blue), with diversity explicitly controlled by a repulsion strength parameter ($\propto\frac{1}{\sigma^2}$), against leading flow-based (purple), score-based (orange), and other non-likelihood methods (black). 
Fitness measures how well generated sequences satisfy the target property (predicted viral packaging efficiency), while diversity quantifies the average Levenshtein distance between sequences in each generated batch. }
\label{fig:aav_inverse_design}
\end{figure}

\paragraph{Protein inverse design.}
In \cref{fig:aav_inverse_design}, we demonstrate our method's performance on the inverse design problem of generating Adeno-Associated Virus (AAV) capsid protein segments \citep{bryant2021deep}. Given a desired functional property (fitness)—here defined as the predicted viral packaging efficiency normalized between 0 and 1—the goal is to design novel protein sequences satisfying this target condition. Beyond achieving high fitness, practical inverse design requires generating diverse candidate sequences to ensure robustness in the downstream experimental validation \citep{jain2022gfnal}. We evaluate on two benchmark splits (\textit{medium} and \textit{hard}), which correspond to subsets of the original AAV dataset differing in baseline fitness distributions and required mutational distance from known high-performing variants \citep{kirjner2023improving}. Leveraging the latent-space representation of VLGPO~\citep{bogensperger2025variational}, we employ our Energy-Matching Langevin sampler with an inference-time tunable repulsion term, allowing explicit control over the diversity of the designed proteins. This enables a flexible trade-off between fitness and diversity, resulting in high fitness scores alongside substantially improved sequence diversity. See \cref{appendix:proteins} for experimental details and dataset descriptions.

\subsection{Local intrinsic dimension estimation} \label{sec:experiments:lid}
Real-world datasets, despite displaying a high number of variables, can often be represented by lower-dimensional manifolds—a concept referred to as the \textit{manifold hypothesis} \citep{fefferman2016testing}. The dimension of such a manifold is called the intrinsic dimension. Estimating the \gls*{lid} at a given point reveals its effective degrees of freedom or \textit{directions of variation}, offering insight into data complexity and adversarial vulnerabilities. We defer the precise definition to \cref{sec:lid_appendix}.

\paragraph{Diffusion-based approaches.}
Recent work leverages pretrained \emph{diffusion models} to estimate the \gls*{lid}~\citep{kamkari2024geometric,stanczuk2024diffusion} by examining the learned score function.
However, since these models do not learn the score at the data manifold (\(t=1\)), their estimates become unreliable there.
Consequently, current methods rely on approximations, for instance by evaluating the score in the proximity of the data manifold (\(t = 1 - t_0\)), where computations remain sufficiently reliable.

\begin{figure}[ht]
    \centering
    \begin{minipage}{.43\linewidth}
        \centering\includegraphics[width=\linewidth]{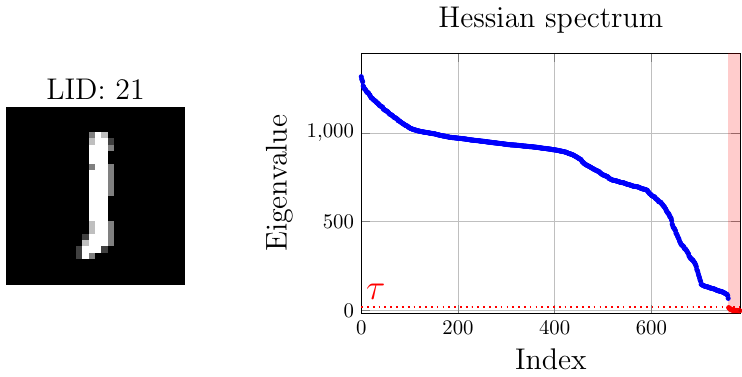}
    \end{minipage}
    \hfill
    \begin{minipage}{.43\linewidth}
        \centering\includegraphics[width=\linewidth]{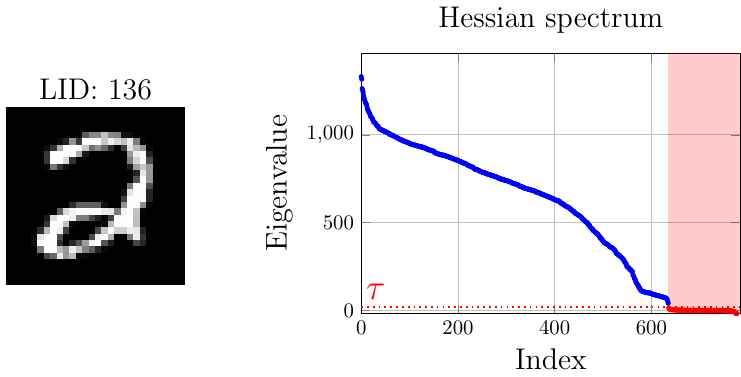}
    \end{minipage}
\caption{Qualitative results for \gls*{lid} estimation using the Hessian spectrum of \(V_\theta(x)\). Left: Spectrum for a low-\gls*{lid} image. Right: Spectrum for a high-\gls*{lid} image. The eigenvalues quantify curvature along principal directions (eigenvectors). A degenerate spectrum (many near-zero eigenvalues, marked in red) indicates locally "flat" regions, revealing the \gls*{lid}. Intuitively, higher image complexity often corresponds to a higher \gls*{lid}.}

    \label{fig:lid}
\end{figure}

\paragraph{Hessian-based LID Estimation.}
Unlike diffusion models, \glspl*{ebm} explicitly parametrize the relative data likelihood.
This explicit parametrization enables efficient analysis of the curvature of the underlying data manifold -- in this example, estimating the \gls*{lid}.
To this end, we compute the Hessian matrix \(\nabla_x^2 V(x_{\mathrm{data}})\) at a given data point and perform its spectral decomposition. We define near-zero eigenvalues as those whose absolute values lie within a small threshold $\tau$ (in our experiments, we set $\tau=3$ for MNIST \citep{deng2012mnist} and $\tau=2$ for CIFAR-10). The count of near-zero eigenvalues reflects the number of \emph{flat} directions and thus reveals the local dimension. As shown in \Cref{tab:lid}, the \gls*{lid} estimates we obtain exhibit stronger correlations with PNG compression size\footnote{PNG is a lossless compression scheme specialized for images and can provide useful guidance when no \gls*{lid} ground truth is available \cite{kamkari2024geometric}.} (evaluated on 4096 images) using Spearman's correlation. \Cref{fig:lid} offers qualitative illustrations. Our \gls*{ebm}-based approach compares favorably to diffusion-based methods, as it relies on fewer approximations by performing computations exactly on the data manifold rather than merely in its vicinity.

\begin{table}
\centering
    \begin{tabular}{c|c|c}
        \cellcolor{black!10}Spearman's correlation $\uparrow$ & \cellcolor{black!10}MNIST & \cellcolor{black!10}CIFAR-10 \\\hline
        ESS \citep{johnsson2014low}& 0.444 & 0.326\\
        FLIPD \citep{kamkari2024geometric} & 0.837 & 0.819\\
        NB \citep{stanczuk2024diffusion}& 0.864 & 0.894\\
        \cellcolor{blue!10} Energy Matching (Ours) & \cellcolor{blue!10}{0.877} & \cellcolor{blue!10}{0.901}\end{tabular}
    \caption{Spearman's correlation coefficients of LID estimates with PNG compression rate. 
 Benchmarks results reported in \citep{kamkari2024geometric}.}
    \label{tab:lid}
\end{table}

\FloatBarrier

\section{Conclusion and limitations}
\label{sec:conclusion}
\paragraph{Contributions.} We introduced a generative framework, \emph{Energy Matching}, that reconciles the advantages of \glspl*{ebm} and \gls*{ot} flow matching models for simulation-free likelihood estimation and efficient high-fidelity generation. Specifically, it: 
\begin{itemize}[leftmargin=*] 
\item Learns a \emph{time-independent scalar potential energy} whose gradient drives rapid high-fidelity sampling--surpassing state-of-the-art energy-based models--while also forming a Boltzmann-like density well suitable for controlled generation. All without auxiliary generators.

\item Offers efficient sampling from target data distributions on par with the state-of-the-art, while learning the score at the data manifold with manageable trainable parameters overhead.

\item Offers a simulation-free, principled likelihood estimation framework for solving inverse problems—where additional priors can be easily introduced—and enables the estimation of a data point’s \gls*{lid} with fewer approximations than score-based methods.
\end{itemize}

\paragraph{Limitations.}
First, our method requires an additional gradient computation with respect to the input, which can increase GPU memory usage (e.g., by 20--40\%), particularly during training. Second, when estimating the \gls*{lid} (Section~\ref{sec:experiments:lid}) for very high-dimensional datasets, computing the full Hessian spectrum may be impractical due to its computational complexity of \( O(d^3) \); in such cases, partial-spectrum methods such as random projections or iterative solvers can be employed instead.

\paragraph{Outlook.} Contrary to widespread belief, we demonstrated that time-independent irrotational methods for generative flows are highly effective and offer an exciting direction for future research. Our Energy Matching approach has the potential to yield novel insights into controlled generation and inverse problems for cancer research \cite{weidner2024learnable,balcerak2024physics}, molecules and proteins \cite{wu2022diffusion,bilodeau2022generative}, computational fluid dynamics \cite{gao2024generative,shysheya2024conditional,molinaro2024generative}, and other fields where precise control over generated samples and effective integration of priors or constraints are crucial. Moreover, Energy Matching aligns naturally with recent generative AI trends toward scaling inference for new capabilities \citep{Zhang_2025_CVPR,Ma2025InferenceTimeScaling}, further broadening its potential impact across scientific and engineering domains.

\begin{ack}
 This research was supported by the Helmut Horten Foundation and the European Cooperation in Science and Technology (COST).
\end{ack}

\bibliographystyle{plainnat}
\bibliography{bib}

\begin{thebibliography}{70}
\providecommand{\natexlab}[1]{#1}
\providecommand{\url}[1]{\texttt{#1}}
\expandafter\ifx\csname urlstyle\endcsname\relax
  \providecommand{\doi}[1]{doi: #1}\else
  \providecommand{\doi}{doi: \begingroup \urlstyle{rm}\Url}\fi

\bibitem[Aggarwal et~al.(2001)Aggarwal, Hinneburg, and Keim]{aggarwal2001surprising}
Charu~C. Aggarwal, Alexander Hinneburg, and Daniel~A. Keim.
\newblock On the surprising behavior of distance metrics in high dimensional space.
\newblock In \emph{International Conference on Database Theory (ICDT)}, 2001.

\bibitem[Albergo and Vanden-Eijnden(2023)]{albergo2023building}
Michael Albergo and Eric Vanden-Eijnden.
\newblock Building normalizing flows with stochastic interpolants.
\newblock In \emph{International Conference on Learning Representations (ICLR)}, 2023.

\bibitem[Ambrosio et~al.(2008)Ambrosio, Gigli, and Savar{\'{e}}]{Ambrosio2008GradientMeasures}
L.~Ambrosio, N.~Gigli, and G.~Savar{\'{e}}.
\newblock \emph{{Gradient flows: in metric spaces and in the space of probability measures}}.
\newblock Lectures in Mathematics ETH Z{\"u}rich. Birkh{\"a}user Basel, 2008.

\bibitem[Ansel et~al.(2024)Ansel, Yang, He, Gimelshein, Jain, Voznesensky, Bao, Bell, Berard, Burovski, et~al.]{ansel2024pytorch}
Jason Ansel, Edward Yang, Horace He, Natalia Gimelshein, Animesh Jain, Michael Voznesensky, Bin Bao, Peter Bell, David Berard, Evgeni Burovski, et~al.
\newblock Pytorch 2: Faster machine learning through dynamic python bytecode transformation and graph compilation.
\newblock In \emph{International Conference on Architectural Support for Programming Languages and Operating Systems (ASPLOS)}, 2024.

\bibitem[Balcerak et~al.(2024)Balcerak, Amiranashvili, Wagner, Weidner, Karnakov, Paetzold, Ezhov, Koumoutsakos, Wiestler, et~al.]{balcerak2024physics}
Michal Balcerak, Tamaz Amiranashvili, Andreas Wagner, Jonas Weidner, Petr Karnakov, Johannes~C Paetzold, Ivan Ezhov, Petros Koumoutsakos, Benedikt Wiestler, et~al.
\newblock Physics-regularized multi-modal image assimilation for brain tumor localization.
\newblock In \emph{Advances in Neural Information Processing Systems (NeurIPS)}, 2024.

\bibitem[Ben-Hamu et~al.(2024)Ben-Hamu, Puny, Gat, Karrer, Singer, and Lipman]{ben2024d}
Heli Ben-Hamu, Omri Puny, Itai Gat, Brian Karrer, Uriel Singer, and Yaron Lipman.
\newblock D-flow: differentiating through flows for controlled generation.
\newblock In \emph{International Conference on Machine Learning (ICML)}, 2024.

\bibitem[Bilodeau et~al.(2022)Bilodeau, Jin, Jaakkola, Barzilay, and Jensen]{bilodeau2022generative}
Camille Bilodeau, Wengong Jin, Tommi Jaakkola, Regina Barzilay, and Klavs~F Jensen.
\newblock Generative models for molecular discovery: Recent advances and challenges.
\newblock \emph{Wiley Interdisciplinary Reviews: Computational Molecular Science}, 12\penalty0 (5):\penalty0 e1608, 2022.

\bibitem[Bogensperger et~al.(2025)Bogensperger, Narnhofer, Allam, Schindler, and Krauthammer]{bogensperger2025variational}
Lea Bogensperger, Dominik Narnhofer, Ahmed Allam, Konrad Schindler, and Michael Krauthammer.
\newblock A variational perspective on generative protein fitness optimization.
\newblock In \emph{International Conference on Machine Learning (ICML)}, 2025.

\bibitem[Brookes et~al.(2019)Brookes, Park, and Listgarten]{brookes2019conditioning}
David Brookes, Hahnbeom Park, and Jennifer Listgarten.
\newblock Conditioning by adaptive sampling for robust design.
\newblock In \emph{International conference on machine learning (ICML)}, 2019.

\bibitem[Bryant et~al.(2021)]{bryant2021deep}
Drew~H. Bryant et~al.
\newblock Deep diversification of an aav capsid protein by machine learning.
\newblock \emph{Nature Biotechnology}, 39:\penalty0 691--696, 2021.

\bibitem[Butcher(2016)]{butcher2016}
John~C. Butcher.
\newblock \emph{Numerical Methods for Ordinary Differential Equations}.
\newblock John Wiley \& Sons, 3rd edition, 2016.

\bibitem[Choi et~al.(2024)Choi, Choi, and Kang]{choi2024scalable}
Jaemoo Choi, Jaewoong Choi, and Myungjoo Kang.
\newblock Scalable wasserstein gradient flow for generative modeling through unbalanced optimal transport.
\newblock In \emph{International Conference on Machine Learning (ICML)}, 2024.

\bibitem[Chrabaszcz et~al.(2017)Chrabaszcz, Loshchilov, and Hutter]{chrabaszcz2017downsampled}
Patryk Chrabaszcz, Ilya Loshchilov, and Frank Hutter.
\newblock A downsampled variant of imagenet as an alternative to the cifar datasets.
\newblock \emph{arXiv preprint arXiv:1707.08819}, 2017.

\bibitem[Chung et~al.(2023)Chung, Kim, Mccann, Klasky, and Ye]{chung2022diffusion}
Hyungjin Chung, Jeongsol Kim, Michael~T Mccann, Marc~L Klasky, and Jong~Chul Ye.
\newblock Diffusion posterior sampling for general noisy inverse problems.
\newblock In \emph{International Conference on Learning Representations (ICLR)}, 2023.

\bibitem[Cui and Han(2024)]{cui2024learning}
Jiali Cui and Tian Han.
\newblock Learning latent space hierarchical ebm diffusion models.
\newblock In \emph{International Conference on Machine Learning (ICML)}, 2024.

\bibitem[Daras et~al.(2024)Daras, Chung, Lai, Mitsufuji, Ye, Milanfar, Dimakis, and Delbracio]{daras2024survey}
Giannis Daras, Hyungjin Chung, Chieh-Hsin Lai, Yuki Mitsufuji, Jong~Chul Ye, Peyman Milanfar, Alexandros~G Dimakis, and Mauricio Delbracio.
\newblock A survey on diffusion models for inverse problems.
\newblock \emph{arXiv preprint arXiv:2410.00083}, 2024.

\bibitem[Deng et~al.(2009)Deng, Dong, Socher, Li, Li, and Fei-Fei]{deng2009imagenet}
Jia Deng, Wei Dong, Richard Socher, Li-Jia Li, Kai Li, and Li~Fei-Fei.
\newblock Imagenet: A large-scale hierarchical image database.
\newblock In \emph{IEEE/CVF Computer Vision and Pattern Recognition (CVPR)}, 2009.

\bibitem[Deng(2012)]{deng2012mnist}
Li~Deng.
\newblock The mnist database of handwritten digit images for machine learning research.
\newblock \emph{IEEE Signal Processing Magazine}, 29\penalty0 (6):\penalty0 141--142, 2012.

\bibitem[Dosovitskiy et~al.(2020)Dosovitskiy, Beyer, Kolesnikov, Weissenborn, Zhai, Unterthiner, Dehghani, Minderer, Heigold, Gelly, et~al.]{dosovitskiy2020image}
Alexey Dosovitskiy, Lucas Beyer, Alexander Kolesnikov, Dirk Weissenborn, Xiaohua Zhai, Thomas Unterthiner, Mostafa Dehghani, Matthias Minderer, G~Heigold, S~Gelly, et~al.
\newblock An image is worth 16x16 words: Transformers for image recognition at scale.
\newblock In \emph{International Conference on Learning Representations (ICML)}, 2020.

\bibitem[Du and Mordatch(2019)]{du2019implicit}
Yilun Du and Igor Mordatch.
\newblock Implicit generation and generalization in energy-based models.
\newblock In \emph{Advances in Neural Information Processing Systems (NeurIPS)}, 2019.

\bibitem[Du et~al.(2021)Du, Li, Tenenbaum, and Mordatch]{du2021improved}
Yilun Du, Shuang Li, Joshua Tenenbaum, and Igor Mordatch.
\newblock Improved contrastive divergence training of energy based models.
\newblock In \emph{International Conference on Machine Learning (ICML)}, 2021.

\bibitem[Fefferman et~al.(2016)Fefferman, Mitter, and Narayanan]{fefferman2016testing}
Charles Fefferman, Sanjoy Mitter, and Hariharan Narayanan.
\newblock Testing the manifold hypothesis.
\newblock \emph{Journal of the American Mathematical Society}, 29\penalty0 (4):\penalty0 983--1049, 2016.

\bibitem[Flamary et~al.(2021)Flamary, Courty, Gramfort, Alaya, Boisbunon, Chambon, Chapel, Corenflos, Fatras, Fournier, et~al.]{Flamary2021}
R{\'e}mi Flamary, Nicolas Courty, Alexandre Gramfort, Mokhtar~Z Alaya, Aur{\'e}lie Boisbunon, Stanislas Chambon, Laetitia Chapel, Adrien Corenflos, Kilian Fatras, Nemo Fournier, et~al.
\newblock {POT: Python Optimal Transport}.
\newblock \emph{Journal of Machine Learning Research}, 22\penalty0 (78):\penalty0 1--8, 2021.

\bibitem[Gao et~al.(2024)Gao, Kaltenbach, and Koumoutsakos]{gao2024generative}
Han Gao, Sebastian Kaltenbach, and Petros Koumoutsakos.
\newblock Generative learning for forecasting the dynamics of high-dimensional complex systems.
\newblock \emph{Nature Communications}, 15\penalty0 (1):\penalty0 8904, 2024.

\bibitem[Gao et~al.(2021)Gao, Song, Poole, Wu, and Kingma]{gaolearning}
Ruiqi Gao, Yang Song, Ben Poole, Ying~Nian Wu, and Diederik~P Kingma.
\newblock Learning energy-based models by diffusion recovery likelihood.
\newblock In \emph{International Conference on Learning Representations (ICLR)}, 2021.

\bibitem[Grathwohl et~al.(2021)Grathwohl, Swersky, Hashemi, Duvenaud, and Maddison]{grathwohl2021oops}
Will Grathwohl, Kevin Swersky, Milad Hashemi, David Duvenaud, and Chris Maddison.
\newblock Oops i took a gradient: Scalable sampling for discrete distributions.
\newblock In \emph{International Conference on Machine Learning (ICML)}, 2021.

\bibitem[Guo et~al.(2023)Guo, Ma, Jiang, Yuan, Yu, and Luo]{guo2023egc}
Qiushan Guo, Chuofan Ma, Yi~Jiang, Zehuan Yuan, Yizhou Yu, and Ping Luo.
\newblock Egc: Image generation and classification via a diffusion energy-based model.
\newblock In \emph{IEEE/CVF International Conference on Computer Vision (ICCV)}, 2023.

\bibitem[Hinton(2002)]{hinton2002training}
Geoffrey~E Hinton.
\newblock Training products of experts by minimizing contrastive divergence.
\newblock \emph{Neural computation}, 14\penalty0 (8):\penalty0 1771--1800, 2002.

\bibitem[Ho et~al.(2020)Ho, Jain, and Abbeel]{ho2020denoising}
Jonathan Ho, Ajay Jain, and Pieter Abbeel.
\newblock Denoising diffusion probabilistic models.
\newblock In \emph{Advances in Neural Information Processing Systems (NeurIPS)}, 2020.

\bibitem[Hopfield(1982)]{hopfield1982neural}
J.~J. Hopfield.
\newblock Neural networks and physical systems with emergent collective computational abilities.
\newblock \emph{Proceedings of the National Academy of Sciences}, 79\penalty0 (8):\penalty0 2554--2558, 1982.

\bibitem[Hyv{\"a}rinen(2006)]{Hyvarinen2006Connections}
Aapo Hyv{\"a}rinen.
\newblock Connections between score matching, contrastive divergence, and pseudolikelihood for continuous-valued data.
\newblock \emph{Neural Computation}, 18\penalty0 (8):\penalty0 1527--1550, 2006.

\bibitem[Ikram et~al.(2024)Ikram, Liu, and Rahman]{ikram2024antibody}
Zarif Ikram, Dianbo Liu, and M~Saifur Rahman.
\newblock Antibody sequence optimization with gradient-guided discrete walk-jump sampling.
\newblock In \emph{ICLR 2024 Workshop on Generative and Experimental Perspectives for Biomolecular Design}, 2024.

\bibitem[Jain et~al.(2022)]{jain2022gfnal}
Moksh Jain et~al.
\newblock Biological sequence design with gflownets.
\newblock In \emph{International Conference on Machine Learning (ICML)}, 2022.

\bibitem[Johnsson et~al.(2014)Johnsson, Soneson, and Fontes]{johnsson2014low}
Kerstin Johnsson, Charlotte Soneson, and Magnus Fontes.
\newblock Low bias local intrinsic dimension estimation from expected simplex skewness.
\newblock \emph{IEEE transactions on pattern analysis and machine intelligence}, 37\penalty0 (1):\penalty0 196--202, 2014.

\bibitem[Jordan et~al.(1998)Jordan, Kinderlehrer, and Otto]{jordan1998variational}
Richard Jordan, David Kinderlehrer, and Felix Otto.
\newblock The variational formulation of the fokker–planck equation.
\newblock \emph{SIAM Journal on Mathematical Analysis}, 29\penalty0 (1):\penalty0 1--17, 1998.

\bibitem[Kamkari et~al.(2024)Kamkari, Ross, Hosseinzadeh, Cresswell, and Loaiza-Ganem]{kamkari2024geometric}
Hamid Kamkari, Brendan Ross, Rasa Hosseinzadeh, Jesse Cresswell, and Gabriel Loaiza-Ganem.
\newblock A geometric view of data complexity: Efficient local intrinsic dimension estimation with diffusion models.
\newblock In \emph{Advances in Neural Information Processing Systems (NeurIPS)}, 2024.

\bibitem[Kim et~al.(2021)Kim, Shin, Song, Kang, and Moon]{kim2021soft}
Dongjun Kim, Seungjae Shin, Kyungwoo Song, Wanmo Kang, and Il-Chul Moon.
\newblock Soft truncation: A universal training technique of score-based diffusion model for high precision score estimation.
\newblock In \emph{International Conference on Machine Learning (ICML)}, 2021.

\bibitem[Kingma and Ba(2014)]{kingma2014adam}
Diederik~P Kingma and Jimmy Ba.
\newblock Adam: A method for stochastic optimization.
\newblock \emph{arXiv preprint arXiv:1412.6980}, 2014.

\bibitem[Kirjner et~al.(2024)]{kirjner2023improving}
Andrew Kirjner et~al.
\newblock Improving protein optimization with smoothed fitness landscapes.
\newblock In \emph{International Conference on Learning Representations (ICLR)}, 2024.

\bibitem[Krizhevsky and Hinton(2009)]{krizhevsky2009learning}
Alex Krizhevsky and Geoffrey~E. Hinton.
\newblock Learning multiple layers of features from tiny images.
\newblock Technical report, University of Toronto, Department of Computer Science, Toronto, Ontario, Canada, 2009.

\bibitem[Lanzetti et~al.(2024)Lanzetti, Terpin, and D{\"o}rfler]{lanzetti2024variational}
Nicolas Lanzetti, Antonio Terpin, and Florian D{\"o}rfler.
\newblock Variational analysis in the wasserstein space.
\newblock \emph{arXiv preprint arXiv:2406.10676}, 2024.

\bibitem[Lanzetti et~al.(2025)Lanzetti, Bolognani, and D{\"o}rfler]{lanzetti2025first}
Nicolas Lanzetti, Saverio Bolognani, and Florian D{\"o}rfler.
\newblock First-order conditions for optimization in the wasserstein space.
\newblock \emph{SIAM Journal on Mathematics of Data Science}, 7\penalty0 (1):\penalty0 274--300, 2025.

\bibitem[LeCun et~al.(2006)LeCun, Chopra, Hadsell, Ranzato, and Huang]{lecun2006tutorial}
Yann LeCun, Sumit Chopra, Raia Hadsell, M~Ranzato, and F~Huang.
\newblock A tutorial on energy-based learning.
\newblock \emph{Predicting Structured Data}, 1\penalty0 (0), 2006.

\bibitem[Lee et~al.(2023)Lee, Jeong, Park, and Shin]{lee2023guiding}
Hankook Lee, Jongheon Jeong, Sejun Park, and Jinwoo Shin.
\newblock Guiding energy-based models via contrastive latent variables.
\newblock \emph{arXiv preprint arXiv:2303.03023}, 2023.

\bibitem[Lee et~al.(2024)Lee, Vecchietti, Jung, Ro, Cha, and Kim]{lee2024robust}
Minji Lee, Luiz~Felipe Vecchietti, Hyunkyu Jung, Hyun~Joo Ro, Meeyoung Cha, and Ho~Min Kim.
\newblock Robust optimization in protein fitness landscapes using reinforcement learning in latent space.
\newblock In \emph{International Conference on Machine Learning (ICML)}, 2024.

\bibitem[Lipman et~al.(2023)Lipman, Chen, Ben-Hamu, Nickel, and Le]{lipmanflow}
Yaron Lipman, Ricky~TQ Chen, Heli Ben-Hamu, Maximilian Nickel, and Matthew Le.
\newblock Flow matching for generative modeling.
\newblock In \emph{International Conference on Learning Representations (ICLR)}, 2023.

\bibitem[Liu et~al.(2023)Liu, Gong, et~al.]{liuflow}
Xingchao Liu, Chengyue Gong, et~al.
\newblock Flow straight and fast: Learning to generate and transfer data with rectified flow.
\newblock In \emph{International Conference on Learning Representations (ICLR)}, 2023.

\bibitem[Liu et~al.(2015)Liu, Luo, Wang, and Tang]{liu2015faceattributes}
Ziwei Liu, Ping Luo, Xiaogang Wang, and Xiaoou Tang.
\newblock Deep learning face attributes in the wild.
\newblock In \emph{International Conference on Computer Vision (ICCV)}, 2015.

\bibitem[Ma et~al.(2025)Ma, Tong, Jia, Hu, Su, Zhang, Yang, Li, Jaakkola, Jia, and Xie]{Ma2025InferenceTimeScaling}
Nanye Ma, Shangyuan Tong, Haolin Jia, Hexiang Hu, Yu‑Chuan Su, Mingda Zhang, Xuan Yang, Yandong Li, Tommi Jaakkola, Xuhui Jia, and Saining Xie.
\newblock Inference‑time scaling for diffusion models beyond scaling denoising steps.
\newblock \emph{arXiv preprint}, arXiv:2501.09732, 2025.

\bibitem[Mardani et~al.(2024)Mardani, Song, Kautz, and Vahdat]{mardani2023variational}
Morteza Mardani, Jiaming Song, Jan Kautz, and Arash Vahdat.
\newblock A variational perspective on solving inverse problems with diffusion models.
\newblock In \emph{International Conference on Learning Representations (ICLR)}, 2024.

\bibitem[Molinaro et~al.(2024)Molinaro, Lanthaler, Raoni{\'c}, Rohner, Armegioiu, Simonis, Grund, Ramic, Wan, Sha, et~al.]{molinaro2024generative}
Roberto Molinaro, Samuel Lanthaler, Bogdan Raoni{\'c}, Tobias Rohner, Victor Armegioiu, Stephan Simonis, Dana Grund, Yannick Ramic, Zhong~Yi Wan, Fei Sha, et~al.
\newblock Generative ai for fast and accurate statistical computation of fluids.
\newblock \emph{arXiv preprint arXiv:2409.18359}, 2024.

\bibitem[Neklyudov et~al.(2023)Neklyudov, Brekelmans, Severo, and Makhzani]{pmlr-v202-neklyudov23a}
Kirill Neklyudov, Rob Brekelmans, Daniel Severo, and Alireza Makhzani.
\newblock Action matching: Learning stochastic dynamics from samples.
\newblock In \emph{International Conference on Machine Learning (ICML)}, 2023.

\bibitem[Shysheya et~al.(2024)Shysheya, Diaconu, Bergamin, Perdikaris, Hern{\'a}ndez-Lobato, Turner, and Mathieu]{shysheya2024conditional}
Aliaksandra Shysheya, Cristiana Diaconu, Federico Bergamin, Paris Perdikaris, Jos{\'e}~Miguel Hern{\'a}ndez-Lobato, Richard Turner, and Emile Mathieu.
\newblock On conditional diffusion models for pde simulations.
\newblock In \emph{Advances in Neural Information Processing Systems (NeurIPS)}, 2024.

\bibitem[Sinai et~al.(2020)Sinai, Wang, Whatley, Slocum, Locane, and Kelsic]{sinai2020adalead}
Sam Sinai, Richard Wang, Alexander Whatley, Stewart Slocum, Elina Locane, and Eric~D Kelsic.
\newblock Adalead: A simple and robust adaptive greedy search algorithm for sequence design.
\newblock \emph{arXiv preprint arXiv:2010.02141}, 2020.

\bibitem[Song and Ermon(2019)]{song2019generative}
Yang Song and Stefano Ermon.
\newblock Generative modeling by estimating gradients of the data distribution.
\newblock In \emph{Advances in neural information processing systems (NeurIPS)}, 2019.

\bibitem[Song and Kingma(2021)]{song2021train}
Yang Song and Diederik~P Kingma.
\newblock How to train your energy-based models.
\newblock \emph{arXiv preprint arXiv:2101.03288}, 2021.

\bibitem[Song et~al.(2021)Song, Sohl-Dickstein, Kingma, Kumar, Ermon, and Poole]{song2021score}
Yang Song, Jascha Sohl-Dickstein, Diederik~P. Kingma, Abhishek Kumar, Stefano Ermon, and Ben Poole.
\newblock Score-based generative modeling through stochastic differential equations.
\newblock In \emph{International Conference on Learning Representations (ICLR)}, 2021.

\bibitem[Stanczuk et~al.(2024)Stanczuk, Batzolis, Deveney, and Sch{\"o}nlieb]{stanczuk2024diffusion}
Jan~Pawel Stanczuk, Georgios Batzolis, Teo Deveney, and Carola-Bibiane Sch{\"o}nlieb.
\newblock Diffusion models encode the intrinsic dimension of data manifolds.
\newblock In \emph{International Conference on Machine Learning (ICML)}, 2024.

\bibitem[Sun et~al.(2025)Sun, Jiang, Zhao, and He]{sun2025noise}
Qiao Sun, Zhicheng Jiang, Hanhong Zhao, and Kaiming He.
\newblock Is noise conditioning necessary for denoising generative models?
\newblock In \emph{International Conference on Machine Learning (ICML)}, 2025.

\bibitem[Terpin et~al.(2024)Terpin, Lanzetti, Gadea, and Dorfler]{terpin2024learning}
Antonio Terpin, Nicolas Lanzetti, Mart{\'\i}n Gadea, and Florian Dorfler.
\newblock Learning diffusion at lightspeed.
\newblock In \emph{Advances in Neural Information Processing Systems (NeurIPS)}, 2024.

\bibitem[Tieleman(2008)]{tieleman2008training}
Tijmen Tieleman.
\newblock Training restricted boltzmann machines using approximations to the likelihood gradient.
\newblock In \emph{International Conference on Machine Learning (ICML)}, 2008.

\bibitem[Tong et~al.(2023)Tong, Fatras, Malkin, Huguet, Zhang, Rector-Brooks, Wolf, and Bengio]{tongimproving}
Alexander Tong, Kilian Fatras, Nikolay Malkin, Guillaume Huguet, Yanlei Zhang, Jarrid Rector-Brooks, Guy Wolf, and Yoshua Bengio.
\newblock Improving and generalizing flow-based generative models with minibatch optimal transport.
\newblock \emph{Transactions on Machine Learning Research}, 2023.

\bibitem[Weidner et~al.(2024)Weidner, Ezhov, Balcerak, Metz, Litvinov, Kaltenbach, Feiner, Lux, Kofler, Lipkova, et~al.]{weidner2024learnable}
Jonas Weidner, Ivan Ezhov, Michal Balcerak, Marie-Christin Metz, Sergey Litvinov, Sebastian Kaltenbach, Leonhard Feiner, Laurin Lux, Florian Kofler, Jana Lipkova, et~al.
\newblock A learnable prior improves inverse tumor growth modeling.
\newblock \emph{IEEE Transactions on Medical Imaging}, 2024.

\bibitem[Welling and Teh(2011)]{welling2011bayesian}
Max Welling and Yee~W Teh.
\newblock Bayesian learning via stochastic gradient langevin dynamics.
\newblock In \emph{International conference on machine learning (ICML)}, 2011.

\bibitem[Wu et~al.(2022)Wu, Gong, Liu, Ye, and Liu]{wu2022diffusion}
Lemeng Wu, Chengyue Gong, Xingchao Liu, Mao Ye, and Qiang Liu.
\newblock Diffusion-based molecule generation with informative prior bridges.
\newblock \emph{Advances in Neural Information Processing Systems (NeurIPS)}, 2022.

\bibitem[Xu et~al.(2023)Xu, Cheng, and Xie]{xu2024normalizing}
Chen Xu, Xiuyuan Cheng, and Yao Xie.
\newblock Normalizing flow neural networks by jko scheme.
\newblock In \emph{Advances in Neural Information Processing Systems (NeurIPS)}, 2023.

\bibitem[Yoon et~al.(2024)Yoon, Hwang, Kwon, Noh, and Park]{yoon2024maximum}
Sangwoong Yoon, Himchan Hwang, Dohyun Kwon, Yung-Kyun Noh, and Frank Park.
\newblock Maximum entropy inverse reinforcement learning of diffusion models with energy-based models.
\newblock In \emph{Advances in Neural Information Processing Systems (NeurIPS)}, 2024.

\bibitem[Zhang et~al.(2025)Zhang, Chu, Berner, Meng, Anandkumar, and Song]{Zhang_2025_CVPR}
Bingliang Zhang, Wenda Chu, Julius Berner, Chenlin Meng, Anima Anandkumar, and Yang Song.
\newblock Improving diffusion inverse problem solving with decoupled noise annealing.
\newblock In \emph{IEEE/CVF Computer Vision and Pattern Recognition (CVPR)}, 2025.

\bibitem[Zhang et~al.(2024)Zhang, Yu, Zhu, Chang, Gao, Wu, and Leong]{zhang2024flow}
Yasi Zhang, Peiyu Yu, Yaxuan Zhu, Yingshan Chang, Feng Gao, Ying~Nian Wu, and Oscar Leong.
\newblock Flow priors for linear inverse problems via iterative corrupted trajectory matching.
\newblock \emph{arXiv preprint arXiv:2405.18816}, 2024.

\bibitem[Zhu et~al.(2024)Zhu, Xie, Wu, and Gao]{zhulearning}
Yaxuan Zhu, Jianwen Xie, Ying~Nian Wu, and Ruiqi Gao.
\newblock Learning energy-based models by cooperative diffusion recovery likelihood.
\newblock In \emph{International Conference on Learning Representations (ICLR)}, 2024.

\end{thebibliography}


\clearpage
\appendix
\section{Additional details on Energy Matching} 
\label{appendix:method-details}
In this section, we provide additional studies and visualizations on our method.

\subsection{Energy landscape during training}
\label{appendix:energy-landscape}
\begin{figure}
    \centering
    \begin{tikzpicture}[>=stealth, node distance=0cm]
        \def\skiptitle{2.0}  
        \def\skipimage{5.0}  
        \def\labelshift{1em}    
        \def\imagewidth{4}      
        \def\imagesep{1em}      

    \node (imgA) [inner sep=0] {
        \includegraphics[width=\imagewidth cm]{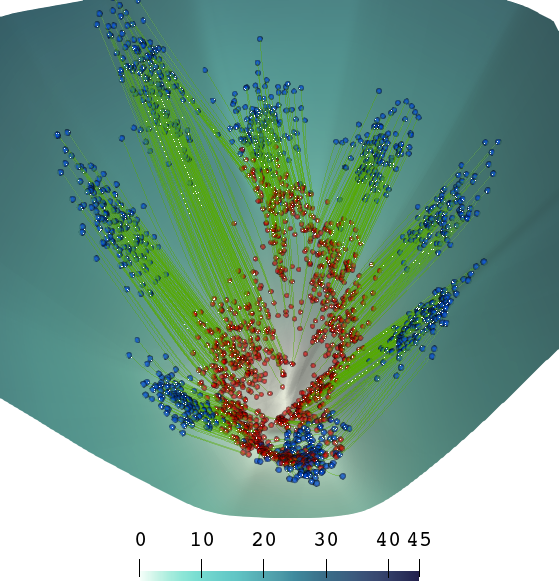}
    };
    \node at ([yshift=\labelshift]imgA.north) {
        \begin{tabular}{c}
            a) Phase 1 (warm-up):\\ $\Lot$
        \end{tabular}
    };
    
    \node (imgB) [inner sep=0, right=\imagesep of imgA] {
        \includegraphics[width=3.78 cm]{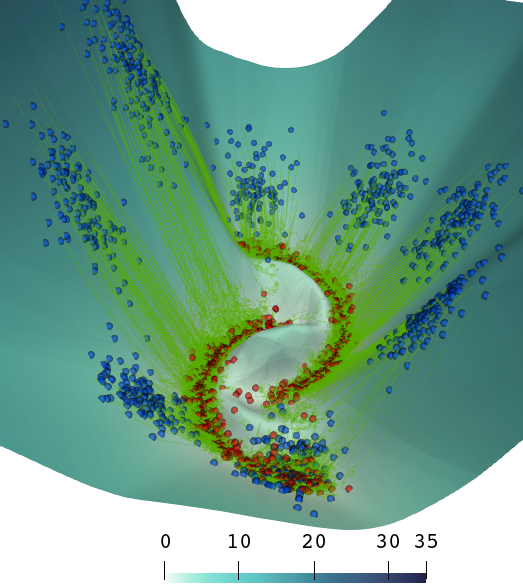}
    };
    \node at ([yshift=\labelshift]imgB.north) {
        \begin{tabular}{c}
            b) Phase 2 (main training):\\ $\Lot+ \Lcd$
        \end{tabular}
    };
    
    \node (imgC) [inner sep=0, right=\imagesep of imgB] {
        \includegraphics[width=\imagewidth cm]{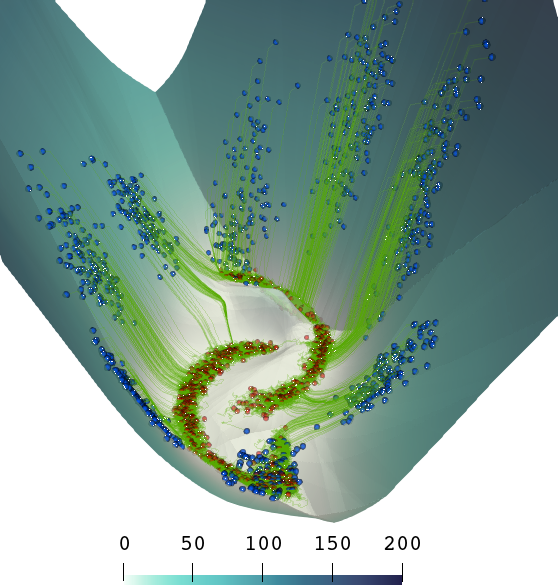}
    };
    \node at ([yshift=\labelshift]imgC.north) {
        \begin{tabular}{c}
            c) Reference EBM:\\ $\Lcd$
        \end{tabular}
    };
    \end{tikzpicture}
\caption{Visualization of the energy $V_\theta(x)$ landscapes driving the samples from eight Gaussians to two moons. See \cref{fig:cover} for the 2D perspective.
         (a) The OT flow loss enforces zero curvature in \(V_{\theta}(x)\) along the trajectories to the target. 
         (b) Around the 2 Moons, the curvature of \(V_{\theta}(x)\) is adjusted to approximate 
         \(\log p_{\text{moons}}(x) \propto V_{\theta}(x)\) while remaining close to the pretrained landscape elsewhere. 
         Combining these objectives yields a potential energy landscape that is both efficient for sampling 
         and representative of the underlying target data distribution.
         (c) An \gls*{ebm} is shown for comparison, trained using contrastive divergence loss. Visible mode collapse that slows down the equilibration. Less regular landscape away from the data as it needs many simulations to explore it.} 
    \label{fig:cover2}
\end{figure}
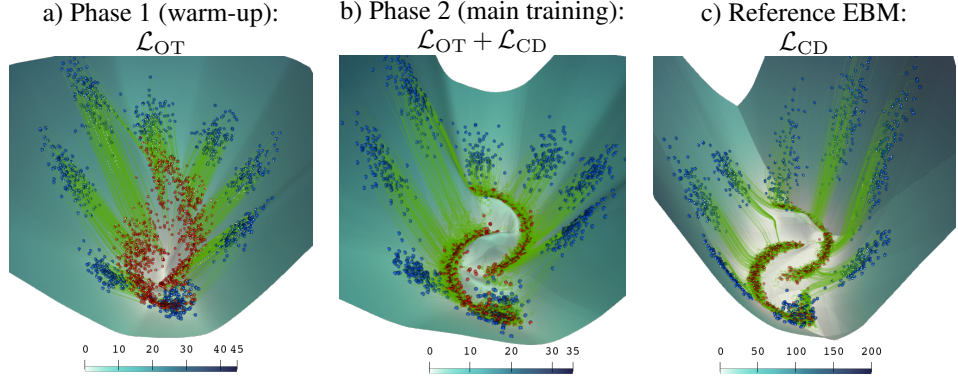

In \cref{fig:cover2}, we visualize how the potential \(V_{\theta}(x)\) transitions from a flow-like regime, where the OT loss enforces nearly zero curvature away from the data manifold (a), to an \gls*{ebm}-like regime, where the curvature around the new data geometry (here, two moons) is adaptively increased to approximate \(\log p_{\text{data}}(x)\) (b). This two-stage design yields a well-shaped landscape that is both efficient to sample (thanks to a mostly flat potential between clusters) and accurate for density estimation near the data modes. For comparison, (c) shows an \gls*{ebm} trained solely with contrastive divergence, exhibiting sharper but less globally consistent basins.

\subsection{Ablation on the sampling time}
\label{appendix:ablations}
\begin{figure}
    \centering
    \begin{tikzpicture}
      \begin{axis}[
        xlabel={Sampling time \(\samplingTime\)},
        ylabel={FID},
        title={FID vs sampling time \(\samplingTime\)},
        xmin=0.7, xmax=3.25,
        ymin=0, ymax=10,
        xtick={1,2,3},
        grid=both,
        axis y discontinuity=crunch,
        legend pos=south east,
        legend style={font=\small},
      ]
    \addplot[
      color=blue,
      mark=*,
      thick,
    ]
    coordinates {
      (0.75, 99.0382)
      (1.00, 14.6512)
      (1.25,  7.4580)
      (1.50,  5.9730)
      (1.75,  5.2009)
      (2.00,  4.6315)
      (2.25,  4.1970)
      (2.50,  3.9051)
      (2.75,  3.6515)
      (3.00,  3.5254)
      (3.25,  3.4151)
    };
    \addlegendentry{\(\Tot=0.9,\ (\Lot+\Lcd)\)}
        
    \addplot[
      color=red,
      mark=*,
      thick,
    ]
    coordinates {
      (0.75, 94.3221)
      (1.00, 13.9849)
      (1.25,  7.1348)
      (1.50,  5.7271)
      (1.75,  4.9980)
      (2.00,  4.4609)
      (2.25,  4.0516)
      (2.50,  3.7784)
      (2.75,  3.5411)
      (3.00,  3.4266)
      (3.25,  3.3270)
    };
    \addlegendentry{\(\Tot=1.0,\ (\Lot+\Lcd)\)}
        
    \addplot[
      color=green!60!black,
      mark=diamond*,
      thick,
    ]
    coordinates {
      (0.75, 98.1233)
      (1.00, 6.6544)
      (1.25, 7.8182)
      (1.50, 11.9746)
      (1.75, 19.2357)
      (2.00, 29.7108)
      (2.25, 42.5727)
      (2.50, 56.1198)
      (2.75, 69.4973)
      (3.00, 81.5454)
      (3.25, 92.2217)
    };
    \addlegendentry{\(\Lot\)}

      \end{axis}
    \end{tikzpicture}
    \caption{CIFAR-10 unconditional generation FID vs.\ sampling time \(\samplingTime\) when sampling from models trained under different scenarios: pure \(\Lot\) and combined \((\Lot+\Lcd)\), with temperature regime switching  parameter \(\Tot \in \{0.9,1.0\}\) during sampling. Lower FID indicates better generative quality. All results for Euler-Heun with $\Delta t = 0.01$.}
    \label{fig:fid_vs_tsample}
\end{figure}
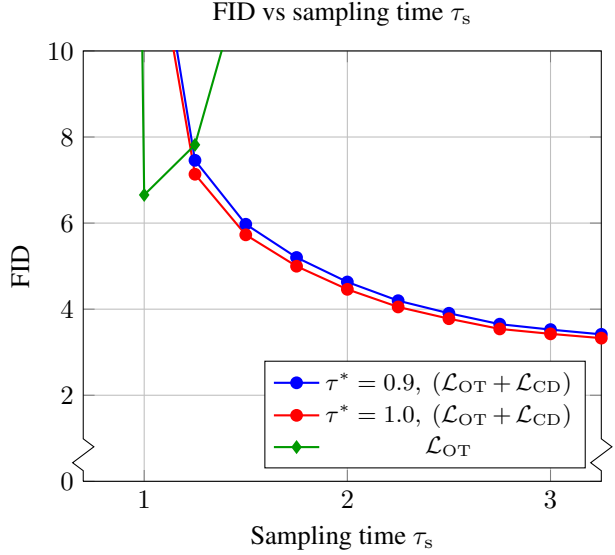

Here, we provide ablation studies on CIFAR-10 unconditional generation. Specifically, we first pretrain using \(\Lot\), and then fine-tune with \((\Lot + \Lcd)\), producing a stable Boltzmann distribution from which one can sample. \cref{fig:fid_vs_tsample} illustrates the FID as a function of sampling time \(\samplingTime\) for models trained under these different regimes. In the case of pure \(\Lot\), the quality measure drops (FID increases) sharply when sampling at \(\samplingTime > 1\); this occurs because, once the samples move close to the data manifold, there is no Boltzmann-like potential well to keep them from drifting away. Because the fidelity slope near the data manifold is steep with respect to sampling time, methods lacking explicit time-conditioning can easily overshoot or undershoot, significantly impacting fidelity. This behavior might explain why some models degrade in performance when made time-independent, as recently reported by \citet{sun2025noise}.

In \cref{fig:fid_vs_tsample} we also report results for different values of the temperature-switching parameter \(\Tot\), which influences the sampling along the paths towards the data manifold (see \cref{eq:eff_temp}). 

\subsection{Ablations on the \gls*{ot} Solver}
\label{app:ot_solver}

We evaluate the computational overhead and sensitivity to solver choice of the \gls*{ot} solver employed in our experiments.

\paragraph{Computational overhead.}  
In the CIFAR-10 experiment, the \gls*{ot} solver accounts for roughly 1.5\% of the training iteration time during Phase~1 \cref{alg:pretrain_revised}. This overhead decreases to a negligible level (approximately 0.01\%) in Phase~2 \cref{alg:fine_tune_revised}, where computational costs are predominantly dominated by the generation of negative samples.

\paragraph{Impact of solver accuracy and complexity.}  
Let us (over‑idealise) and model a standardised CIFAR‑10 image as a vector \( x \in \mathbb{R}^d \), drawn from \(\mathcal{N}(0, I_d)\), with dimensionality \( d = 32 \times 32 \times 3 = 3072 \), and paired with an independent Gaussian-noise vector \( z \sim \mathcal{N}(0, I_d) \). Each coordinate of the difference \( (z - x) \) thus follows \( \mathcal{N}(0, 2) \), and the squared Euclidean distance distribution is \( |z - x|_2^2 \sim 2\chi_d^2 \), which has mean \(2d\), standard deviation \(\sqrt{8d}\), and relative spread \(\frac{\sqrt{8d}}{2d} \approx 0.025\). This demonstrates the "thin-shell" phenomenon, implying that all entries of the cost matrix \( C_{ij} = |z_i - x_j|^2_2 \) concentrate around nearly identical values, consistent with the distance-concentration effect observed by \citep{aggarwal2001surprising}. Consequently, the choice among exact linear programming (LP), entropic regularisation (Sinkhorn), or even random matching should yield nearly identical cumulative optimal-transport costs, despite their different complexities: \(O(n^3 \log n)\) for LP, \(O(\frac{n^2}{\kappa})\) for Sinkhorn (with regularisation strength \(\kappa\)), and \(O(n)\) for random matching.

Empirical evidence supporting this is summarised below:
\begin{itemize}
    \item Our CIFAR-10 runs: FID degrades from 3.34 (LP) to 3.37 (random).
    \item \citep{tongimproving} Table~5: FID scores of 4.44 (LP) vs 4.46 (random) at 100 steps.
    \item \citep{tongimproving} Fig.~D.2: solver accuracy saturates beyond batch size 16 in the 2D task.
    \item \citep{terpin2024learning} App.~C.2: LP and Sinkhorn methods produce indistinguishable results unless regularisation is extreme.
\end{itemize}

We employ the LP approach for robustness, negligible cost, and no additional hyperparameters.

\subsection{Sampling Time Analysis}
\label{appendix:sampling_time}

Computing the gradient \(\nabla_x V_{\theta}(x)\) via automatic differentiation (\texttt{autograd} \citep{ansel2024pytorch}) introduces additional computational overhead compared to directly evaluating \(V_{\theta}(x)\). Specifically, on the CIFAR-10 network architecture (see \cref{fig:cifar_diagram}), gradient evaluation is approximately \(2.15\times\) slower, as it requires both forward and backward passes. In contrast, flow matching and diffusion models directly parameterize the velocity field, thus only requiring forward computations during sampling.

Nevertheless, despite this per-step computational cost, our method achieves competitive overall sampling efficiency due to a reduced number of integration steps needed for high-quality generation. As demonstrated in \cref{tab:sampling_time_analysis}, Energy Matching achieves a lower FID (3.34) in 173 seconds per batch, outperforming OT-FM (FID 3.74 in 136 seconds per batch) and DDPM++ (FID 3.45 in 183 seconds per batch). Our results thus indicate a favorable balance between computational overhead per step and total sampling runtime.

\begin{table*}[!h]
    \centering
    \caption{Comparison of sampling efficiency and quality on CIFAR-10 (batch size 128, NVIDIA R6000 48GB GPU). Despite gradient computation overhead (\(\nabla_x V_\theta(x)\) via backward pass), Energy Matching achieves superior FID scores with competitive wall-clock sampling time.}
    \label{tab:sampling_time_analysis}
    \small
    \begin{tabular}{lcccc}
        \toprule
        \textbf{Method} & \textbf{Params} & \textbf{Steps} & \textbf{Sampling Time [s]↓} & \textbf{FID↓} \\
        \midrule
        \textbf{Flow-/Diffusion-based Models} & & & & \\
        \quad OT-FM \citep{tongimproving} & 37M & 1000 & \textbf{136} & 3.74 \\
        \quad DDPM++ \citep{kim2021soft}  & 62M & 1000 & 183 & 3.45 \\
        \midrule
        \textbf{Energy-based Models} & & & & \\
        \rowcolor{blue!10}
        \quad Energy Matching \textit{(Ours)} & 50M & 325 & 173 & \textbf{3.34} \\
        \bottomrule
    \end{tabular}
\end{table*}

\section{Details on AAV inverse design protein generation}
\label{appendix:proteins}

We optimize protein fitness for adeno-associated virus (AAV) sequences in the \textit{medium} and \textit{hard} data regimes proposed by \citep{kirjner2023improving}, using latent encodings and backbone architectures from \citep{bogensperger2025variational}. Conditional sampling employs classifier guidance via learned predictor networks \(g_{\phi}\) or \(\tilde g_{\phi}\) to steer samples toward high-fitness regions. The CNN-based fitness predictors from \citep{kirjner2023improving} are trained only on the limited training data for each regime.

Training follows \cref{alg:pretrain_revised} and \cref{alg:fine_tune_revised}. We sample 128 sequences using \cref{alg:sampling_inverse_interaction}, keeping the same batch size across all baselines; detailed hyperparameters are given in \cref{appendix:training-details}. Generated sequences are evaluated for fitness using the learned oracle from \citep{kirjner2023improving}, and further assessed for both intra-set diversity and novelty relative to the training sequences \citep{jain2022gfnal}. Our approach achieves state-of-the-art fitness while improving diversity (see \cref{table:aav_results}). Incorporating interaction energy in \cref{alg:sampling_inverse_interaction} further enhances diversity with manageable impact on fitness.

\begin{table*}[h]
    \centering
    \caption{AAV optimization results. For VLGPO (flow-based) and Energy Matching, medium difficulty uses \(g_{\phi}\), hard difficulty uses \(\tilde g_{\phi}\). Metrics (\textit{Fitness}↑, \textit{Diversity}↑, \textit{Novelty}↑). Reported uncertainty of Fitness is expressed as standard deviation.}
    \label{table:aav_results}
    \scriptsize
    \begin{tabular}{l ccc ccc}
        \toprule
         & \multicolumn{3}{c}{\textbf{AAV medium}} & \multicolumn{3}{c}{\textbf{AAV hard}} \\
        \cmidrule(lr){2-4} \cmidrule(lr){5-7}
        \textbf{Method} & \textbf{Fitness↑} & \textbf{Diversity↑} & \textbf{Novelty↑} & \textbf{Fitness↑} & \textbf{Diversity↑} & \textbf{Novelty↑} \\
        \midrule
        \multicolumn{7}{l}{\textbf{Learning Unnormalized Data Likelihood}} \\
        \quad\textbf{Energy-based Models} \\
        \rowcolor{blue!10}
        \quad Energy Matching \textit{(\textbf{Ours})}                   & 0.59 (0.0) & 5.86 & 5.0  & 0.61 (0.0) & 4.77  & 6.7  \\
        \rowcolor{blue!10}
        \quad Energy Matching (+repulsion) \textit{(\textbf{Ours})}     & 0.58 (0.0) & 6.22 & 5.0  & 0.60 (0.0) & 5.22  & 6.6  \\
        \midrule
        \multicolumn{7}{l}{\textbf{Learning Transport/Score Along Noised Trajectories}} \\
        \quad\textbf{Flow-based Models} \\
        \quad VLGPO \citep{bogensperger2025variational}    & 0.58 (0.0) & 5.58 & 5.0  & 0.61 (0.0) & 4.29  & 6.2  \\
        \midrule
        \quad\textbf{Diffusion Models} \\
        \quad gg-dWJS \citep{ikram2024antibody}            & 0.48 (0.0) & 9.48 & 4.2  & 0.33 (0.0) & 14.3  & 5.3  \\
        \midrule
        \textbf{Other Methods} \\
        \quad LatProtRL \citep{lee2024robust}               & 0.57 (0.0) & 3.00 & 5.0  & 0.57 (0.0) & 3.00  & 5.0  \\
        \quad GGS \citep{kirjner2023improving}              & 0.51 (0.0) & 4.00 & 5.4  & 0.60 (0.0) & 4.50  & 7.0  \\
        \quad AdaLead \citep{sinai2020adalead}             & 0.46 (0.0) & 8.50 & 2.8  & 0.40 (0.0) & 8.53  & 3.4  \\
        \quad CbAS \citep{brookes2019conditioning}         & 0.43 (0.0) & 12.70& 7.2  & 0.36 (0.0) & 14.4  & 8.6  \\
        \quad GWG \citep{grathwohl2021oops}                & 0.43 (0.1) & 6.60 & 7.7  & 0.33 (0.0) & 12.0  & 12.2 \\
        \quad GFN-AL \citep{jain2022gfnal}                                      & 0.20 (0.1) & 9.60 & 19.4 & 0.10 (0.1) & 11.6  & 19.6 \\ 
        \bottomrule
    \end{tabular}
\end{table*}
\FloatBarrier

\section{Details on LID estimation}
\label{sec:lid_appendix}

\paragraph{Definition.}
To start, we need to introduce the concept of \emph{local mass}, defined as
\[
M(r) \;=\; \int_{B(x_\text{data};r)} p(x)\,dx,
\]
where \(p(x)\) is the local density and $B(x_\text{data}, r)$ is a ball of radius $r$ around $x_\text{data}$, i.e. $B(x_\text{data}, r) = \{x \in \mathbb{R}^d : \|x - x_{\mathrm{data}}\|\le r\}$. The \gls*{lid} is then given by:
\[
\mathrm{LID}(x_{\mathrm{data}})
= d- 
\lim_{r \to 0}
\frac{\log\bigl(M(r)\bigr)}{\log(r)}.
\]
Intuitively, \(M(r)\) measures how much probability mass is concentrated in a ball of radius \(r\) around \(x_{\mathrm{data}}\). As we shrink this ball, the growth rate of \(M(r)\) in terms of \(r\) reveals the local dimensional structure of the data.

\paragraph{Assumptions.}
In the context of contrastive divergence, we assume that data points \(x_{\mathrm{data}}\) lie in well-like regions \citep{Hyvarinen2006Connections}, i.e.:
\[
\nabla V(x_{\mathrm{data}}) \approx 0 
\quad\text{and}\quad
\nabla^2 V(x_{\mathrm{data}}) \text{ is positive semidefinite (or nearly so).}
\]
Conceptually, \(V(x)\) can be thought of as an energy function; points where \(\nabla V(x_{\mathrm{data}}) = 0\) are near local minima of this energy, and the Hessian \(\nabla^2 V(x_{\mathrm{data}})\) provides information about local curvature (see \cref{fig:lid} for a qualitative illustration).

\paragraph{Energy-based density.}
We define an energy-based density
\[
p(x) \;\propto\; \exp\!\Bigl(-\tfrac{V(x)}{\varepsilon}\Bigr),
\]
where \(\varepsilon\) is a temperature parameter. Near a data point \(x_{\mathrm{data}}\) satisfying \(\nabla_x V(x_{\mathrm{data}})=0\), we can approximate \(V(x)\) by its second-order Taylor expansion:
\[
V(x) \approx
V(x_{\mathrm{data}}) + 
\frac{1}{2} (x - x_{\mathrm{data}})^\top \nabla_x^2 V(x_{\mathrm{data}}) (x - x_{\mathrm{data}}).
\]
Consequently, in view of the assumptions above,
\[
p(x) 
\propto
\exp\left(
  -\frac{1}{2\varepsilon}
  (x - x_{\mathrm{data}})^\top \nabla_x^2 V(x_{\mathrm{data}})(x - x_{\mathrm{data}})
\right).
\]

\paragraph{Local mass derivation and the rank of the energy Hessian.}
Substituting the local quadratic form of $p(x)$ near $x_{\mathrm{data}}$ into the definition of the local mass $M(r)$, we obtain:
\[
M(r) 
= 
\int_{B(x_\text{data}, r)} 
p(x)dx
\propto
\int_{B(x_\text{data}, r)}
\exp\left(
  -\tfrac{1}{2\varepsilon}
  (x - x_{\mathrm{data}})^\top\nabla_x^2 V(x_{\mathrm{data}})(x - x_{\mathrm{data}})
\right) dx.
\]
For small $r$, the dominant contribution depends on the rank of the Hessian $\nabla_x^2 V(x_{\mathrm{data}})$. Let $k = \mathrm{rank}(\nabla_x^2 V(x_{\mathrm{data}}))$. Then, as $r \to 0$, one can show that $
M(r) = Cr^k$,
where $C$ does not depend on $r$. We take the 
logarithm on both sides and divide by $\log(r)$ to get
\[
\frac{\log(M(r))}{\log(r)} = \frac{\log(C) + k \log(r)}{\log(r)} = k + \frac{\log(C)}{\log(r)},
\]
and the second term vanishes as \( r \to 0 \). Hence,
\[
\mathrm{LID}(x_{\mathrm{data}})=d-k.
\]

\paragraph{Practical estimation.}
In practice, the \gls*{lid} at a data point \( x_{\mathrm{data}} \) can be estimated through the following procedure:
\begin{enumerate}[leftmargin=*]
  \item Train $V(x)$ with Energy Matching.
  \item Compute the Hessian \( H = \nabla_x^2 V(x_{\mathrm{data}}) \).
  \item Perform an eigenvalue decomposition on \( H \).
\end{enumerate}
Then the estimated local data-manifold dimension corresponds to the number of directions with negligible curvature (smaller magnitude than some $\tau$).
\section{Training details} \label{appendix:training-details}
Below, we detail the training configurations for CIFAR-10, ImageNet 32x32, CelebA, MNIST, and AAV. Additionally, we provide intuitions for practical hyperparameter choices to facilitate effective training across additional datasets. We recommend using SiLU activation functions wherever possible, as they smooth out the energy landscape and improve the numerical stability of the $\nabla_x V(x)$ computation. The gradient of the potential, $\nabla_x V(x)$, is computed using automatic differentiation via PyTorch's \texttt{autograd} \citep{ansel2024pytorch}. We optimize all models using the Adam optimizer \citep{kingma2014adam} and maintain an exponential moving average (EMA) of the model weights.

While we specifically adopt
(i) a one-sided trimmed mean of negative sample energies and
(ii) clamping of the contrastive loss for stability,
any commonly used EBM technique (e.g., persistent contrastive divergence~\citep{tieleman2008training},
replay buffers, multi-scale negative sampling) could be readily employed.

In our approach, we introduce two hyperparameters, \(\alpha\) and \(\beta\), to control these stabilizing techniques:
\[
\begin{aligned}
\alpha &= \text{fraction of negative energies discarded to remove outliers that skew the mean (e.g., top 10\%)},\\
\beta &= \text{clamp threshold for $\mathcal{L}_{\mathrm{CD}}$ (i.e., we clamp $\mathcal{L}_{\mathrm{CD}}$ to be $\geq -\beta$).}
\end{aligned}
\]

\begin{figure}[ht]
\centering
\resizebox{0.9\textwidth}{!}{
\begin{minipage}{0.47\textwidth}
\centering
\begin{tikzpicture}[font=\small, >=latex,
  every node/.style={draw, rectangle, rounded corners=2pt, align=center, minimum height=0.9cm, minimum width=2.2cm}]
  \node (input) at (0,0) {Input\\(3$\times$32$\times$32)};
  \node (unet) [below=0.7cm of input] {UNet\\(37M params)};
  \node (transformer) [below=0.7cm of unet] {Transformer\\(PatchEmbed+ViT, 12M params)};
  \node (output) [below=0.7cm of transformer] {Scalar\\Potential};
  
  \draw[->, thick] (input) -- (unet);
  \draw[->, thick] (unet) -- (transformer);
  \draw[->, thick] (transformer) -- (output);
\end{tikzpicture}
\end{minipage}%
\hfill
\begin{minipage}{0.5\textwidth}
\centering
\small
\begin{tabular}{ll}
\hline
\textbf{Hyperparameter} & \textbf{Value} \\
\hline
Image size & 3$\times$32$\times$32 \\
Base channels (UNet) & 128 \\
ResBlocks & 2 \\
channel\_mult & [1, 2, 2, 2] \\
Attention resolution & 16 \\
Attention heads (UNet) & 4 \\
Head channels (UNet) & 64 \\
Dropout & 0.1 \\
\hline
\multicolumn{2}{l}{\textit{Transformer (ViT) Head}} \\
\hline
Patch size & 4 \\
Embedding dim & 384 \\
Transformer layers & 8 \\
Transformer heads & 4 \\
Output scale & 1000.0 \\
\hline
\end{tabular}
\end{minipage}
}
\caption{\small Diagram of our UNet+Transformer EBM for CIFAR-10 and ImageNet 32x32. A UNet (37M params) processes a 3$\times$32$\times$32 image; its output is fed into a Transformer head (PatchEmbed + 8-layer ViT, 12M params) that produces a scalar potential. Here we employ the identical UNet architecture as in \citep{tongimproving}, but with the time parameter fixed at $t=0$ to render the model time-independent.}
\label{fig:cifar_diagram}
\end{figure}
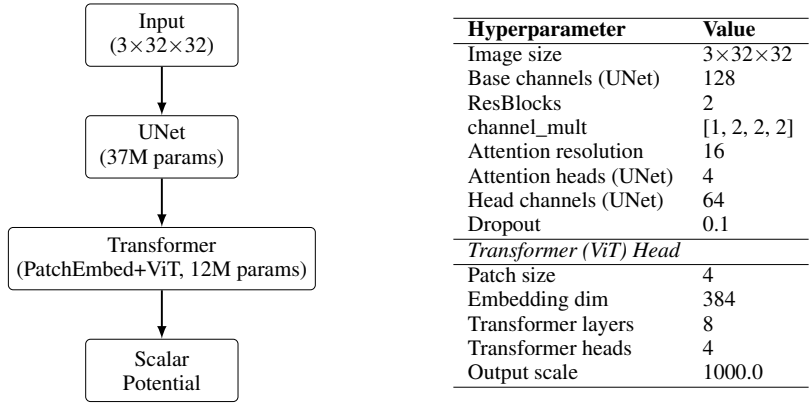

\paragraph{CIFAR-10:} The architecture is shown in \cref{fig:cifar_diagram}. We use the same UNet from \citep{tongimproving}  (with fixed $t=0.0$, making it effectively time-independent) followed by a small vision transformer (ViT) \citep{dosovitskiy2020image} to obtain a scalar output. Hyperparameters are: $\tau_s = 3.25$, $\tau^* = 1.0$, $\Delta t = 0.01$, $M_{\text{Langevin}} = 200$. We train for 145k iterations using \cref{alg:pretrain_revised} with EMA $0.9999$ and then 2k more with \cref{alg:fine_tune_revised} and  EMA $0.99$ on 4xA100. The batch size is 128, learning rate is $1.2\times10^{-3}$, $\varepsilon_{\max}=0.01$,  $\lambda_{\mathrm{CD}}=1\times10^{-3}$, $\alpha = 0.1$, and $\beta=0.02$. Negatives initialized on the data manifold follow the same temperature schedule as those initialized from the noise.

\paragraph{ImageNet 32x32:} The architecture is shown in \cref{fig:cifar_diagram} (same as for CIFAR-10). Hyperparameters are: $\tau_s = 2.5$, $\tau^* = 1.0$, $\Delta t = 0.01$, $M_{\text{Langevin}} = 200$. We train for 640k iterations using \cref{alg:pretrain_revised} with EMA $0.9999$ and then 1k more with \cref{alg:fine_tune_revised} and EMA $0.99$ on 7xA100. The batch size is 128, learning rate is $6\times10^{-4}$, $\varepsilon_{\max}=0.01$, $\lambda_{\mathrm{CD}}=1\times10^{-3}$, $\alpha = 0.1$, and $\beta=0.02$. Negatives initialized on the data manifold follow the same temperature schedule as those initialized from the noise.

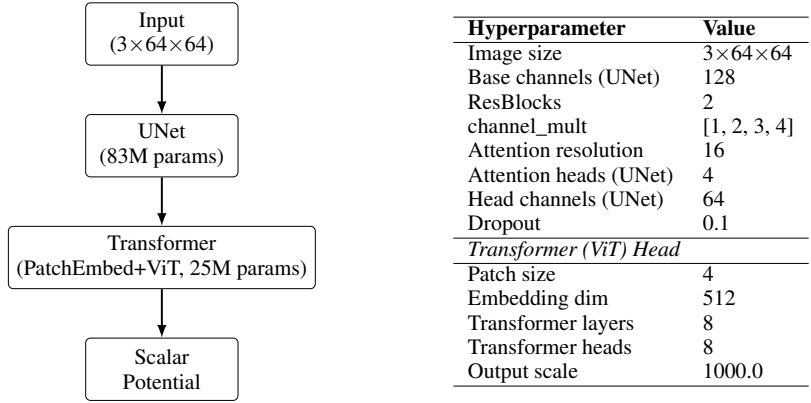
\begin{figure}[ht]
\centering
\resizebox{0.9\textwidth}{!}{
\begin{minipage}{0.47\textwidth}
\centering
\begin{tikzpicture}[font=\small, >=latex,
  every node/.style={draw, rectangle, rounded corners=2pt, align=center, minimum height=0.9cm, minimum width=2.2cm}]
  \node (input) at (0,0) {Input\\(3$\times$64$\times$64)};
  \node (unet) [below=0.7cm of input] {UNet\\(83M params)};
  \node (transformer) [below=0.7cm of unet] {Transformer\\(PatchEmbed+ViT, 25M params)};
  \node (output) [below=0.7cm of transformer] {Scalar\\Potential};
  
  \draw[->, thick] (input) -- (unet);
  \draw[->, thick] (unet) -- (transformer);
  \draw[->, thick] (transformer) -- (output);
\end{tikzpicture}
\end{minipage}%
\hfill
\begin{minipage}{0.5\textwidth}
\centering
\small
\begin{tabular}{ll}
\hline
\textbf{Hyperparameter} & \textbf{Value} \\
\hline
Image size & 3$\times$64$\times$64 \\
Base channels (UNet) & 128 \\
ResBlocks & 2 \\
channel\_mult & [1, 2, 3, 4] \\
Attention resolution & 16 \\
Attention heads (UNet) & 4 \\
Head channels (UNet) & 64 \\
Dropout & 0.1 \\
\hline
\multicolumn{2}{l}{\textit{Transformer (ViT) Head}} \\
\hline
Patch size & 4 \\
Embedding dim & 512 \\
Transformer layers & 8 \\
Transformer heads & 8 \\
Output scale & 1000.0 \\
\hline
\end{tabular}
\end{minipage}
}
\caption{\small Diagram of our UNet+Transformer EBM for CelebA. A UNet (83M params) processes a 3$\times$64$\times$64 image; its output is fed into a Transformer head (PatchEmbed+8-layer ViT, 25M params) that produces a scalar potential.}
\label{fig:celeba_diagram}
\end{figure}

\paragraph{CelebA:} We scale the CIFAR-10 model by $\sim2\times$; see \cref{fig:celeba_diagram}. We set $\tau_s = 2.0$, $\tau^* = 1.0$, $\Delta t = 0.01$, $M_{\text{Langevin}} = 200$, and train for 250k iterations using \cref{alg:pretrain_revised} with EMA $0.9999$ then 4k with \cref{alg:fine_tune_revised} and EMA $0.99$ on 4xA100. The batch size is 32, learning rate is $1\times10^{-4}$, $\varepsilon_{\max}=0.05$, $\lambda_{\mathrm{CD}}=1\times10^{-4}$.

\paragraph{MNIST:} We downscale the CIFAR-10 model (\cref{fig:cifar_diagram}) to 2M parameters by reducing the UNet base width to 32 channels, using channel multipliers [1, 2, 2], setting the number of attention heads in the UNet to 2, simplifying the Transformer head to an embedding dimension of 128, 2 Transformer layers, 2 attention heads, and adjusting the output scale to 100.0. We set $\tau_s = 2.0$, $\tau^* = 1.0$, $\Delta t = 0.025$, $M_{\text{Langevin}} = 75$, and train for 50k iterations using \cref{alg:pretrain_revised} with EMA $0.999$ then 3.3k with \cref{alg:fine_tune_revised} and EMA $0.99$ on a single A100. The batch size is 128, learning rate is $1\times10^{-4}$, $\varepsilon_{\max}=0.1$, $\lambda_{\mathrm{CD}}=1\times10^{-3}$, $\alpha = 0.0$, and $\beta=0.05$.  Negatives initialized on the data manifold follow the same temperature schedule as those initialized from the noise.

\paragraph{AAV:} We adopt the one-dimensional CNN architecture as used in \citep{bogensperger2025variational}, summing the final-layer activations to obtain the potential. We train for 10k iterations using \cref{alg:pretrain_revised} and for 1k iterations using \cref{alg:fine_tune_revised} on a single A100. The batch size is 128, learning rate is $1\times10^{-4}$, $\varepsilon_{\max}=0.1$, $M_{\text{Langevin}} = 200$, $\Delta t = 0.01$, and $\lambda_{\mathrm{CD}}=1\times10^{-4}$. For \cref{alg:sampling_inverse_interaction} we use $\tau_s = 1.7$ for AAV medium and $\tau_s = 1.3$ for AAV hard, $\tau^* = 0.9$, $\zeta=0.01$ for AAV medium and $\zeta=0.009$ for AAV hard. We set the target fitness to $y=1$ to aim for the maximum fitness in the generated sequences.

\paragraph{Intuition for other datasets:}
It is essential for negative samples to reach the equilibrium distribution induced by the model or at least the proximity of the data manifold. The condition $M_{\text{Langevin}} \times \Delta t \gg 1$ is critical to achieving this, with $\Delta t$ small enough to ensure that negative samples remain of sufficient quality—typically the same $\Delta t$ as used in flow matching for the given generation task. In practice, we set $M_{\text{Langevin}} \times \Delta t = 2$ across most experiments. We recommend  starting with $\tau^{*} = 1.0$ to ensure optimal transport regularization near the data manifold, thereby enhancing training stability. If additional conditions are required during sampling, exploring lower values ($\tau^{*} < 1.0$) may be beneficial, as this parameter does not need to remain consistent between training and sampling (as shown in \cref{fig:fid_vs_tsample}). Training with $\tau^{*} < 1.0$ is advised only in special cases, such as extremely low-dimensional problems like that shown in \cref{fig:cover}, where it is possible to simultaneously be far from the data manifold (from the perspective of the target mode) and close to it (from the perspective of another mode). The parameter $\varepsilon_{\max}$ controls how extensively negative samples explore the space. For unconditional generation, we use $\varepsilon_{\max}=0.01$, but for inverse problems or design tasks, higher values (e.g., $\varepsilon_{\max}=0.05$) can improve robustness. The parameter $\tau_s$ significantly depends on the task (unconditional or conditional) and thus must be tuned accordingly post-training. Without explicit tuning, selecting $\tau_s = M_{\text{Langevin}} \times \Delta t$ is a reasonable default. Finally, parameters $\lambda_{\mathrm{CD}}$, $\alpha$, and $\beta$ influence the stability of contrastive training and must be empirically determined alongside appropriate early stopping.

\end{document}